\documentclass[lettersize,journal]{IEEEtran}

\usepackage{amsmath,amsfonts,amssymb}
\usepackage{amsthm}
\usepackage{array}
\usepackage{booktabs}
\usepackage{textcomp}
\usepackage{stfloats}
\usepackage{url}
\usepackage{graphicx}
\usepackage{cite}
\usepackage{multirow}
\usepackage{enumitem}
\newtheorem{proposition}{Proposition}

\newtheorem{remark}{Remark}

\providecommand{\Rp}{\mathbb{R}_+}
\providecommand{\dd}{\,\mathrm{d}}

\begin{document}

\title{Partially Observed Sparse Graphs:\\ The Unknown Sampling Rate is a Tail Index}

\author{Jian~Xu,
        Delu~Zeng,
        John~Paisley,
        and~Qibin~Zhao%
\thanks{J. Xu is with RIKEN iTHEMS, Wako, Japan, and with the RIKEN Center for
Advanced Intelligence Project (AIP), Tokyo, Japan
(e-mail: jian.xu@riken.jp).}%
\thanks{D. Zeng is with the South China University of Technology, Guangzhou,
China.}%
\thanks{J. Paisley is with Columbia University, New York, NY, USA.}%
\thanks{Q. Zhao is with the RIKEN Center for Advanced Intelligence Project (AIP),
Tokyo, Japan.}%
}

\maketitle

\begin{abstract}
A large graph is often available only in part: a crawl stopped by its budget, a
panel, a partial dump. When the sampled fraction $s$ is known by design the total
edge count follows from $\hat e=e_s/s^2$ and no model is needed. We treat the case
where $s$ is unknown and the population size is known. Our main result is a
reduction: under a sparse exchangeable (graphex) model the expected non-isolated
fraction obeys $n_s/n_1\to s^{1+\sigma}$, so the sampling rate becomes estimable once
the tail index $\sigma$ is, and substituting it back gives $e_s(n_1/n_s)^{2/(1+\sigma)}$
--- the same estimator, with the design quantity inferred. Estimating global edge
cardinality in a sparse graph is therefore, in expectation, tail-index estimation,
and the quadratic graphon estimator is the case $\sigma=0$: it fails by an identity
rather than by a fit ($260\%$ median error against $27\%$). We bound the finite-size
error of the substitution and show the reduction is \emph{modular} in the tail-index
estimator --- filled with a published closed-form one it reaches $21.7\%$ over $13$
networks and $39$ sampling budgets with no fitting at all. Fitting a full graphex
additionally returns the degree distribution at any size and a generative object, in
a representation where sparsity is a coordinate and the interpolation path is
dictated rather than chosen. Two limits are exact: rank-one graphexes have
transitivity fixed by the degree profile, so high-clustering graphs lie outside the
class; and under snowball or random-walk crawls every method here fails, the
design-based oracle worst of all ($7.8\%$ to $588\%$).
\end{abstract}


\section{Introduction}
\label{sec:intro}

\IEEEPARstart{L}{arge} graphs are routinely available only through a sample. A crawler
exhausts its budget after reaching a fraction of the vertices; a platform exposes a
panel rather than its full user base; a measurement infrastructure observes a
network too large to enumerate. In each case one holds a subgraph and wants global
properties of the graph it came from: the total number of edges, how the degree
distribution continues past what was seen, how the two scale against one another as
the observed fraction grows.

For sparse graphs this is harder than it first appears, and the difficulty is
specific. The quantity that governs the answer is the \emph{densification exponent}
$\beta$ relating edge count to node count, $e\asymp n^{\beta}$; it is not a
statistic of the sample but a property of the whole, and estimating it wrongly costs
a factor $R^{\Delta\beta}$ at extrapolation ratio $R$. Assuming no densification
($\beta=1$) is badly wrong on real networks, and assuming the dense-limit value
($\beta=2$) is worse. Throughout this paper the observed graph is a
uniform $p$-sample of the target and the target is the graph it was drawn from. Two
things are therefore outside our scope, and we mark them here rather than in a
limitations section: predicting a separately collected or future network, which is a
different problem; and degree-biased exploration --- snowball and random-walk crawls
--- under which every method we test, including the design-based oracle, fails by
margins we measure in Section~\ref{sec:crawl}.

Graphons are the canonical limits of \emph{dense} sequences. Real networks are not
dense: citation, autonomous-system and peer-to-peer graphs have edge counts close
to linear in the node count. This is not a minor mismatch. For fixed $W$, sampling
$N$ latent positions gives $\mathbb{E}[e]=\tfrac{N^2}{2}\iint W$, so the edge count
is $\Theta(N^2)$ \emph{by construction}. Fitting such a model at $n$ nodes to a
family whose law is $e\asymp n$ and sampling at $n'$ over-predicts edges by $n'/n$.
This is an identity; no training or architecture changes it. \cite{rozada2026diphon}
state the limitation themselves and list ``extending the formal analysis beyond the
dense regime'' as future work.

The limit theory for sparse graphs exists. Sparse exchangeable graphs and their
limits, \emph{graphexes}, were developed by \cite{caron2017sparse},
\cite{veitch2015class} and \cite{borgs2018sparse}: a graphex is sampled from a
Poisson process on $\Rp^2$ restricted to a size parameter, is projective by
construction, and admits sparsity exponents strictly between linear and quadratic.
To our knowledge it has no deep generative counterpart.

What a graphex buys here is a \emph{reduction}. The obstacle above is that $\beta$ is
not observable; the observation is that it does not have to be, because the same
model that makes $\beta$ meaningful also ties the unknown sampling rate to the
observable node counts, and the two facts cancel. This is the paper's main claim, and
everything else is subordinate to it: which estimator fills the resulting slot is a
separate and replaceable question, and the generative half is a second use of the
same representation, in which the interpolation path turns out to be determined by
the parameterisation rather than free. Both halves come with a ceiling we can state
exactly.

\paragraph{Contributions.}
\begin{enumerate}[topsep=2pt,itemsep=1pt]
\item \textbf{A statistical reduction for unknown-$s$ partial graph estimation.} The
  design-based estimator $\hat e=e_s/s^2$, the natural answer when the inclusion
  probability is known, remains available when it is not: under a graphex the
  retained-vertex fraction obeys $n_s/n_1\to s^{1+\sigma}$, so $s$ is recoverable from
  observable node counts once the tail index $\sigma=1/\alpha$ is, and
  $e_s(n_1/n_s)^{2/(1+\sigma)}$ is \emph{the same estimator}
  (Section~\ref{sec:design}). This, not any particular estimator of $\sigma$, is what
  the paper claims. The quadratic graphon estimator is the special case $\sigma=0$,
  which is why it fails by an identity rather than by a fitting error.
\item \textbf{The reduction is modular, and its modularity has a limit.} Any
  $\hat\sigma$ can fill the slot; filled with the closed-form estimator of
  \cite{naulet2021bootstrap} it reaches $21.7\%$ median error with no fitting.
  Substituting the same $\hat\sigma$ into the degree-profile fit, however, makes the
  degree distribution worse: the extreme tail governs the scaling exponent and the
  bulk governs the histogram, and a single object must trade them off
  (Section~\ref{sec:modular}).
\item \textbf{A finite-size deviation bound.} The reduction is a limit statement about
  expectations, applied to one finite graph. Proposition~\ref{prop:conc} bounds the
  variances of both counts by Poisson-functional arguments and shows the resulting
  relative errors vanish, leaving the regular-variation remainder as the one
  uncontrolled term (Section~\ref{sec:finite}).
\item \textbf{A parametric graphex fit, and the flow over it.} Deriving a closed-form
  expression for the expected degree counts and fitting the resulting model
  numerically yields the degree distribution at any size and a generative object, not
  only an exponent; we compare it against \cite{naulet2021bootstrap}, against an
  adapted form of the degree-distribution inverse estimator of
  \cite{zhang2015estimating}, and against simpler alternatives, and report where each
  wins --- which for both tasks is not us. Over those objects, in the canonical
  parameterisation we adopt, the interpolation path is not free: arithmetic
  interpolation makes $\alpha_t$ constant on the interior and discontinuous at the
  endpoint (Proposition~\ref{prop:arith}), whereas geometric interpolation keeps the
  family closed with $\alpha_t$ linear (Proposition~\ref{prop:geom}) and is linear in
  log-space, so the model is ordinary CFM.
\item \textbf{Two relations that bound what is achievable}: an error-propagation
  relation from tail-index error to edge-count error (Proposition~\ref{prop:prop},
  verified at $r=0.978$), and a clustering ceiling for rank-one graphexes
  (Proposition~\ref{prop:trans}) that our generation experiments hit exactly.
\end{enumerate}

\paragraph{What is established empirically, and what is not.} The \emph{framework}
--- that global edge estimation under partial observation reduces to recovering the
sampling rate through the tail index --- is what the real-network experiments test,
and it is where the large effects are: any exponent below $2$ removes most of the
graphon's error and the isolate correction accounts for the rest. Within it we do
\emph{not} claim our tail-index estimator is the most accurate available. The
published graphex estimator of \cite{naulet2021bootstrap} is at least as good for
edge count and an adapted form of \cite{zhang2015estimating} is ahead of us for the
degree distribution (Sections~\ref{sec:baselines} and~\ref{sec:degdist}); our fit
earns its place by delivering exponent, distribution at any size and generator from
one object, not by winning either comparison. The \emph{flow} contributes that
construction, with a path fixed by Proposition~\ref{prop:geom} rather than chosen,
and its evidence is synthetic: the real-network experiments test the scaling law and
the estimator under the $p$-sampling protocol above, not the generative component,
and we do not read them as evidence for it. A reader interested only in predicting
how a sparse network densifies needs Sections~\ref{sec:theory}
and~\ref{sec:method}; the flow matters if one wants to sample graphs rather than
exponents.

\paragraph{Why a flow model at all?} Since the estimation results do not depend on
it, the question deserves a direct answer rather than an implicit one. The flow is
not needed to predict an edge count, and we do not claim it is. It is needed when
the output must be a \emph{graph} rather than a number --- releasing a synthetic
stand-in for a network that cannot be shared, generating load for a system sized to
a population one has only sampled, simulating a process on graphs drawn from a
fitted population rather than on one point estimate. For those uses the object to be
learned is a distribution over graphexes, and Proposition~\ref{prop:geom} is then not
a modelling choice but a constraint: on graphex space the interpolation path
determines whether the quantity that carries sparsity is even present along the
trajectory, and this holds for any generative model that interpolates in this
parameterisation, not only for conditional flow matching. What we establish about
the flow is established on synthetic data, where ground truth is analytic; we say so
again in Section~\ref{sec:gen} rather than letting the real-network numbers stand in
for it.

\section{Background}
\label{sec:bg}

\paragraph{Graphex processes.} A graphex is a triple $(I,S,W)$ with $W:\Rp^2\to[0,1]$
symmetric measurable; $I$ and $S$ carry isolated edges and stars and play no role
below, so we write $W$. Sampling at size $\nu$ takes a unit-rate Poisson process on
$[0,\nu]\times\Rp$ with points $(\theta_i,\vartheta_i)$ and connects $i\sim j$ with
probability $W(\vartheta_i,\vartheta_j)$. Since $\theta$ does not affect
connectivity, the sample at $\nu'<\nu$ is the induced subgraph on
$\{\theta_i\le\nu'\}$ (\emph{projectivity}); and because $\theta$ is uniform and
independent, restricting to $\theta\le s\nu$ is distributionally identical to
retaining each node independently with probability $s$. We use this throughout:
\textbf{uniform random node sub-sampling of an observed graph, followed by
discarding the vertices left isolated, is exactly graphex sub-sampling} --- the
discard step is needed because the observable graph is by definition the
non-isolated part. This makes our estimators implementable on real data, where
$\theta$ is unobservable.

\paragraph{Flow matching.} Conditional flow matching \cite{lipman2022flow, liu2022flow, albergo2022building} learns a velocity field whose ODE transports a
source to a target by regressing on a conditional velocity along a prescribed path;
for a linear path this is $x_1-x_0$. Section~\ref{sec:theory} shows that on graphex
space the path cannot be chosen freely.

\section{Related work}
\label{sec:related}

\paragraph{Graph sampling and property estimation.} Estimating properties of a
network from a sample is a long-established problem.
\cite{leskovec2006sampling} compare sampling designs by how well the sample
preserves a battery of structural statistics, and \cite{ahmed2013network} extend
the comparison to streaming settings; \cite{kolaczyk2009statistical} gives the
statistical framing and \cite{liu2018graph} surveys the neighbouring problem of
compressing a graph while preserving its statistics. \cite{jiao2024sampling} studies
the design question in the regime this paper works in, where the sampling rate is low.
A separate line builds design-based estimators for specific functionals: random-walk
and multidimensional-walk crawlers with Horvitz--Thompson-style corrections
\cite{ribeiro2010estimating, gjoka2010walking}, network size
\cite{katzir2011estimating}, average degree \cite{dasgupta2014estimating}, graphlet
statistics from a walk \cite{chen2016general}, and transitivity and triangle counts
from a stream in small space \cite{jha2013space, zhang2023counting} --- the last being
the closest existing analogue of a global statistic that our
Proposition~\ref{prop:trans} shows is \emph{not} free to vary once the degree profile
is fixed. Estimation from a deliberately restricted view has also been studied for
privacy rather than for budget \cite{liu2024edge}, and sampling as a scalability
device inside a learned model is a line of its own \cite{ding2025scalable}.
This literature has also documented repeatedly that
samples misrepresent tails: subgraphs of scale-free graphs are not scale-free
\cite{stumpf2005subnets}, and traceroute-style exploration manufactures power laws
in graphs that have none \cite{achlioptas2009bias}. Our target --- a global count
at a size larger than anything observed --- differs from the usual one of
estimating a functional at the observed size, and it is what forces an exponent
into the problem.

\paragraph{Degree-distribution reconstruction.} \cite{zhang2015estimating} pose
recovery of the full degree distribution from a sampled network as an ill-posed
inverse problem and solve it with a penalised weighted least-squares estimator.
That work recovers the distribution at the \emph{observed} population size; we fit a
parametric tail so the same object can be evaluated at a larger size
(Section~\ref{sec:degdist}).

\paragraph{Sparse exchangeable graphs.} Graphex processes
\cite{caron2017sparse, veitch2015class, borgs2018sparse} give projective models with
sparsity strictly between linear and quadratic; \cite{veitch2019sampling} develop
the sampling and estimation theory, and edge-exchangeable constructions
\cite{crane2018edge, cai2016edge} reach sparsity by a different route. Most directly
related to us, \cite{naulet2021bootstrap} give a nonparametric estimator of the
graphex tail index from the observed degree counts, with a consistency theorem and a
rate.

\paragraph{What is new relative to \cite{naulet2021bootstrap} and
\cite{zhang2015estimating}.} Since both are ahead of our estimator on their own task,
the difference has to be stated at the level of theorems rather than of numbers.
\cite{naulet2021bootstrap} estimate $\sigma$ from one observed graph and stop there;
\cite{zhang2015estimating} reconstruct a degree distribution at the size at which it
was observed. Neither addresses a quantity at a size larger than the sample, and
neither treats the sampling rate as an unknown to be recovered --- \cite{zhang2015estimating}
require the design to be known and \cite{naulet2021bootstrap} sub-sample the observed
graph at rates they choose. Our Section~\ref{sec:design} is the step neither takes:
it identifies $s$ itself as the estimand, shows that the graphex scaling law makes it
identifiable from observable counts, and derives that the resulting plug-in
\emph{is} the design-based estimator, which turns a modelling question into an
identification one and explains the graphon's failure as the case $\sigma=0$.
Propositions~\ref{prop:conc} and~\ref{prop:remainder} then bound the two errors this
substitution introduces at finite size. Their estimators fill a slot inside that
statement rather than competing with it, which is why
Section~\ref{sec:modular} recommends \cite{naulet2021bootstrap} for the slot.

\paragraph{Deep generative graph models.} Discrete diffusion
\cite{vignac2022digress} and discrete flow matching dominate graph generation (see
\cite{cao2024survey} for the wider diffusion literature), with recent work on few-step
sampling \cite{roos2026categorical}, on maintaining hard structural constraints along
the trajectory \cite{madeira2024generative}, and on unifying generation with
prediction in a latent space \cite{zhou2024unifying}. Scale has been attacked directly
by exploiting sparsity inside the denoiser \cite{qin2023sparse}, which lowers the cost
of representing a large graph but leaves the size at which it is generated an input
rather than a modelled quantity. These operate at a node count supplied at sampling
time and do not address the scaling law of the generated family.
On size generalisation, \cite{rozada2026diphon} define a diffusion on graphon space
so that one object can be discretised at any size, and \cite{bergmeister2024efficient}
reach several thousand nodes by iterative local expansion; the former's guarantees
are dense-regime, which its authors name as a limitation, and the latter has no limit
object or explicit sparsity exponent. Both fix the edge count to be quadratic in the
node count. We are orthogonal to both: we identify which quantity must be
extrapolated and make it a parameter, and we are not aware of a flow or diffusion
model defined on graphex space.

\section{Theory}
\label{sec:theory}

Write $\mu(x)=\int W(x,y)\dd y$ for the marginal.

\subsection{The scaling law}

\begin{proposition}[Scaling]
\label{prop:scaling}
Let $\iint W<\infty$ and $\mu(x)\sim c\,x^{-\alpha}$ as $x\to\infty$, $\alpha>1$. Then
\begin{align}
e_\nu &= \frac{\nu^2}{2}\iint W = \Theta(\nu^2), \label{eq:enu}\\
n_\nu &= \nu\!\int_0^\infty\!\bigl(1-e^{-\nu\mu(x)}\bigr)\dd x \nonumber\\
      &= \nu(\nu c)^{1/\alpha}\Gamma\!\left(1-\tfrac1\alpha\right)(1+o(1)),
\label{eq:nnu}
\end{align}
hence $\mathbb{E}[e]\asymp\mathbb{E}[n]^{\beta}$ with
$\beta=\dfrac{2\alpha}{\alpha+1}\in(1,2)$. All scaling statements in this paper are
at the level of expectations; we do not prove concentration, and the sample-level
relation $e\asymp n^{\beta}$ should be read as shorthand for the expectation-level
one.
\end{proposition}

The correspondence between the tail of $\mu$ and the sparsity exponent is part of
the sparse-graphex literature \cite{caron2017sparse, veitch2015class, borgs2018sparse, veitch2019sampling}; we restate it in this form because the constant and the
$\Gamma$-factor are what the estimator of Section~\ref{sec:method} inverts, and
because the two consequences below are what the rest of the paper rests on.

First, $e_\nu$ is $\Theta(\nu^2)$
\emph{whatever the shape of $W$}: community or block structure does not affect the
scaling, which is carried entirely by the tail of $\mu$. Second, $\alpha\to1^+$
gives $\beta\to1$ (genuinely sparse) and $\alpha\to\infty$ gives $\beta\to2$; the
graphon case is the boundary $\beta=2$. We therefore write
\begin{equation}
\label{eq:decomp}
W(x,y) = \underbrace{\psi(x)\psi(y)}_{\text{degree profile: sparsity}}
\cdot\underbrace{K(x,y)}_{\text{shape: structure}},
\end{equation}
with $\psi$ regularly varying of index $\alpha$ and $K$ bounded, tending to $1$ in
the tail.

\subsection{Tail extrapolation is unavoidable}
\label{sec:tail}

Sampling at $\nu'\gg\nu$ places points at values of $\vartheta$ never populated at
training size, so a model that is nonparametric everywhere says nothing about the
region governing the extrapolation. This is a statement about the observation
process, not about an architecture. We keep the bulk of $\psi$ and all of $K$
nonparametric and give the tail a regularly-varying form with few parameters.

\subsection{Why the path must be geometric}

Let $\alpha_0<\alpha_1$, so $\mathcal{W}_0$ has the heavier tail.

\begin{proposition}[Arithmetic interpolation destroys the exponent]
\label{prop:arith}
For $W_t=(1-t)W_0+tW_1$ we have $\mu_t=(1-t)\mu_0+t\mu_1$, and since the tail of a
sum is dominated by the heavier summand, $\alpha_t=\alpha_0$ for every $t\in[0,1)$,
with $\alpha_1$ attained only at $t=1$. The map $t\mapsto\alpha_t$ is constant on
the interior and discontinuous at the endpoint.
\end{proposition}

Two readings must be separated here. Arithmetic interpolation does not annihilate
all information about $\alpha_1$ --- the lighter-tailed term is still present, merely
sub-leading as $x\to\infty$. What it does is leave the single-power-law family in
which we have chosen to represent the tail, so that the \emph{leading tail index},
the coordinate the model transports, is uninformative along the interior. This is a
closure argument about the selected representation, not a general impossibility
result for arithmetic paths on graphex space.

\begin{proposition}[Geometric interpolation is closed, with a linear exponent]
\label{prop:geom}
For $W_t=W_0^{1-t}W_1^{t}$: (i) $W_t$ is symmetric, measurable, $[0,1]$-valued and
integrable, since $a^{1-t}b^{t}\le(1-t)a+tb$ by weighted AM--GM; (ii) if both endpoints are written in
the same ordering coordinate (Remark~\ref{rem:invariance}) with
$\psi_i(x)=c_i(1+x)^{-\alpha_i}$ on the tail, then
$\psi_t(x)=c_0^{1-t}c_1^{t}(1+x)^{-[(1-t)\alpha_0+t\alpha_1]}$, so
$\alpha_t=(1-t)\alpha_0+t\alpha_1$; (iii) the decomposition \eqref{eq:decomp} is
preserved with $K_t=K_0^{1-t}K_1^{t}$; (iv) the conditional velocity is
$\partial_t W_t=W_t\log(W_1/W_0)$ whenever $W_0>0$.
\end{proposition}

Part (ii) is the technical anchor: the parametric tail family is \emph{closed}
under the geometric path, which is exactly what arithmetic interpolation cannot
provide, since a sum of power laws is not a power law and $\alpha_t$ is then not
even defined within the family.

\begin{remark}[Which parts of the path are representation-independent]
\label{rem:invariance}
A graphex is defined only up to measure-preserving relabelings: $W$ and
$W\circ(\varphi\times\varphi)$ describe the same object. Pointwise interpolation is
\emph{not} invariant under this equivalence --- replacing $W_1$ by a relabeled
representative changes $W_0^{1-t}W_1^{t}$ --- so a path defined by
Proposition~\ref{prop:geom} presupposes a choice of representative. Three things
follow, and we separate them because the first is often mistaken for the second.

\emph{The tail index of a single graphex is an invariant.} A measure-preserving
relabeling permutes $\mu$ but leaves its decreasing rearrangement unchanged, so
$\alpha$ is a well-defined function on the equivalence class, and the statements
that concern one graphex at a time --- Propositions~\ref{prop:scaling}
and~\ref{prop:prop} --- are representation independent.

\emph{The tail index along a path between two graphexes is not.} It would be wrong
to conclude from the previous paragraph that
$\alpha_t=(1-t)\alpha_0+t\alpha_1$ holds for any pair of representatives.
Invariance of $\alpha_0$ and of $\alpha_1$ separately says nothing about the
pointwise product: for a relabeling $\varphi$ the path passes through
$\psi_0(x)^{1-t}\psi_1(\varphi(x))^{t}$, whose decreasing rearrangement depends on
how the two profiles are aligned, exactly as the law of a product of two random
variables depends on their coupling and not only on the two marginals. Nothing
prevents $\varphi$ from placing the light part of $\psi_1$ where $\psi_0$ is heavy,
and the resulting exponent is then not a convex combination of the endpoints'.
Proposition~\ref{prop:geom}(ii) is proved by multiplying two power laws at the same
argument, which presupposes that both endpoints are written in the same ordering
coordinate.

\emph{What we therefore claim.} We define the path on a canonical representative,
not on the equivalence class: our parameterisation constrains $\lambda$ to be
non-increasing, i.e.\ it works with the decreasing rearrangement of the degree
profile, and assigns positions by inverting the cumulative count
(Section~\ref{sec:method}), so the representative is determined by the data up to
ties. Propositions~\ref{prop:arith} and~\ref{prop:geom} are statements about that
canonical parameterisation: geometric interpolation is closed under it and
arithmetic interpolation is not. They are not statements that geometric
interpolation is forced on graphex space as a quotient, and we do not make that
claim. The distinction matters for how much the result proves, and we prefer the
weaker reading to an overstated one.

Choosing the representative this way is the one-dimensional analogue of the
alignment problem that makes quotient constructions hard on labelled graphs, and it
is easy here for a specific reason: sorting a marginal is $O(n\log n)$, whereas
aligning two adjacency matrices is a quadratic assignment problem. We do not claim
the path is canonical beyond this ordering.
\end{remark}

\begin{remark}[The model is ordinary CFM]
\label{rem:cfm}
With $\ell=\log W$, $\lambda=\log\psi$, $\kappa=\log K$, the geometric path is the
\emph{linear} path $\ell_t=(1-t)\ell_0+t\ell_1$ with conditional velocity
$\ell_1-\ell_0$. The constraint $\ell\le0$ is preserved by convexity, so no
projection is needed; $\alpha$ enters $\lambda$ linearly and is transported as an
ordinary coordinate. The construction is standard CFM on a finite vector.
\end{remark}

\subsection{A ceiling on clustering in the rank-one case}

The following is stated and proved for rank-one kernels; Section~\ref{sec:gen}
reports what we observe for the kernels our estimator actually produces.

\begin{proposition}[Clustering is fixed by the degree profile]
\label{prop:trans}
Let $W=\psi(x)\psi(y)$ with $\int\psi<\infty$ and $\int\psi^{2}<\infty$, and let
edges be independent Bernoulli with $p_{ij}=\psi_i\psi_j\le1$. Write
$S=\sum_i\psi_i$, $S_2=\sum_i\psi_i^{2}$, and let $T$ and $P$ be the numbers of
triangles and of connected triples. Then $\mathbb{E}[T]=S_2^3/6$ and
$\mathbb{E}[P]=S^2S_2/2$ to leading order, so the \emph{expected-count
transitivity}
\begin{align}
\bar\tau \;:=\; \frac{3\,\mathbb{E}[T]}{\mathbb{E}[P]}
&=\frac{3\cdot S_2^3/6}{S^2S_2/2}=\left(\frac{S_2}{S}\right)^{\!2}\nonumber\\
&\longrightarrow\;\left(\frac{\int\psi^2}{\int\psi}\right)^{\!2}
\label{eq:ceiling}
\end{align}
is \textbf{independent of the size parameter $\nu$} and determined by the degree
profile alone.
\end{proposition}

\paragraph{What is and is not asserted.} The statement is about the ratio of
expected counts, $3\mathbb{E}[T]/\mathbb{E}[P]$, not about the expectation of the
random ratio $3T/P$ that an empirical transitivity measures; the two agree only
under a concentration argument we do not give, so \eqref{eq:ceiling} should be read
as a statement about the population quantity. Two remarks make this less of a gap
than it may appear. Both $T$ and $P$ are sums of weakly dependent indicators and
$\mathbb{E}[P]\to\infty$ in the regime of interest, which is the usual setting in
which such ratios converge; and the discrepancy is measurable, so we measure it.
Section~\ref{sec:verify} finds the ratio of empirical to predicted transitivity
falling from $1.13$ at $n=300$ to $1.05$ at $n=1000$ and $1.04$ at $n=3000$ ---
converging as the counts grow, which is the behaviour a concentration argument
would establish. We therefore use
\eqref{eq:ceiling} as a ceiling on what the construction can reach, and verify it
numerically rather than claiming it as an almost-sure limit.

\paragraph{Scope of this result, stated carefully.} The proposition is proved for
the \emph{rank-one} case $W=\psi(x)\psi(y)$ only. Our model is $W=\psi\psi K$ with a
non-trivial $K$ on the bulk, and we do \emph{not} claim the ceiling for general $K$:
for a general kernel the triangle integral depends on $K$, and matching the degree
marginal does not by itself pin the triangle density. Two things are nonetheless
true of our parameterisation. It is exactly rank-one outside the bulk cut-off
($K\equiv1$ for $\max(x,y)>R$), so at large $\nu$, where the tail carries almost all
vertices, the proposition applies to an increasing fraction of the graph. And
empirically the ceiling is not relieved by giving $K$ more capacity: increasing $R$
sevenfold moves the generated transitivity from $0.032$ to $0.024$ against a true
$0.037$ and never changes the sign of the assortativity error
(Section~\ref{sec:ablation}). Whether some admissible $K$ within this class could
reach high clustering is left open; what we establish is the rank-one identity, and
what we observe is that our estimated $K$ does not escape it.

\subsection{From tail-index error to edge-count error}

\begin{proposition}[Error propagation]
\label{prop:prop}
Let $\hat\alpha$ estimate $\alpha$ and let $R$ be the extrapolation ratio in node
count. Since $\beta=2\alpha/(\alpha+1)$ has $\dd\beta/\dd\alpha=2/(\alpha+1)^2$,
\begin{align}
\left|\frac{\hat e-e}{e}\right| &= \left|R^{\hat\beta-\beta}-1\right|\nonumber\\
&\approx\;|\hat\alpha-\alpha|\cdot\frac{2\ln R}{(\alpha+1)^2}
\;+\;\text{(sampling floor)} ,
\label{eq:prop}
\end{align}
where the floor is the Poisson variability of the observed counts and does not
vanish with $|\hat\alpha-\alpha|$.
\end{proposition}

This converts a target edge accuracy at a given extrapolation ratio into a required
estimator accuracy, and is verified in Section~\ref{sec:verify}.

\subsection{A finite-size deviation bound}
\label{sec:finite}

Propositions~\ref{prop:scaling} and~\ref{prop:prop} are statements about
expectations, while the estimator is applied to one finite graph. The gap is not
rhetorical: \eqref{eq:shat} substitutes a limiting ratio for an observed one, and
nothing so far says the observed ratio is close to it. The following controls the
stochastic half of that gap.

\begin{proposition}[Relative concentration of the counts]
\label{prop:conc}
Let $W$ be a graphex with $\iint W<\infty$ and $\int\mu^2<\infty$, and let $n_\nu$
and $e_\nu$ be the numbers of non-isolated vertices and of edges at size $\nu$. Then
\begin{align}
\mathrm{Var}(e_\nu) &\le 3e_\nu + \nu^3\!\int\!\mu^2 ,\label{eq:vare}\\
\mathrm{Var}(n_\nu) &\le 12e_\nu + 4\nu^3\!\int\!\mu^2 .\label{eq:varn}
\end{align}
If in addition $\mu(x)\sim cx^{-\alpha}$ with $\alpha>1$, then
\begin{equation}
\frac{\mathrm{sd}(e_\nu)}{\mathbb{E}[e_\nu]}=O(\nu^{-1/2}),
\qquad
\frac{\mathrm{sd}(n_\nu)}{\mathbb{E}[n_\nu]}=O\!\left(\nu^{\frac12-\frac1\alpha}\right).
\label{eq:rates}
\end{equation}
The edge count therefore concentrates relatively for every $\alpha>1$; the bound for
the vertex count vanishes if and only if $\alpha<2$.
\end{proposition}

The proof is in Appendix~\ref{app:proofs}: a Poincar\'e inequality
functionals applied to the conditional mean, plus an Efron--Stein bound for the edge
coins. The step that keeps the bound finite is that adding a point changes the vertex
count only through the vertices it connects to, so the difference operator is
controlled by that point's degree rather than by a constant.

Two consequences. First, combining \eqref{eq:rates} with \eqref{eq:shat}, the
observed ratio satisfies $n_{s\nu}/n_\nu=(\mathbb{E}n_{s\nu}/\mathbb{E}n_\nu)
(1+O_P((s\nu)^{1/2-1/\alpha}))$, so the relative error of the plug-in decomposes as
\begin{align}
\left|\frac{\hat e - e}{e}\right| \;\lesssim\;
&\underbrace{\frac{2}{1+\sigma}\,\delta_\nu}_{\text{regular-variation remainder}}
\;+\;\underbrace{O_P\!\left((s\nu)^{\frac12-\frac1\alpha}\right)}_{\text{sampling}}
\nonumber\\[2pt]
&\;+\;\underbrace{|\hat\alpha-\alpha|\,\frac{2\ln R}{(\alpha+1)^2}}
   _{\text{Proposition~\ref{prop:prop}}} .
\label{eq:decomp3}
\end{align}
Proposition~\ref{prop:conc} controls the middle term. The first, $\delta_\nu$, is the
deterministic error in replacing $\mathbb{E}n_{s\nu}/\mathbb{E}n_\nu$ by
$s^{1+\sigma}$, and a second-order condition makes it explicit.

\begin{proposition}[Rate of the regular-variation remainder]
\label{prop:remainder}
Suppose $\mu$ is bounded on $[0,x_0]$ and there are $c>0$, $\rho>0$, $C<\infty$ with
$|\mu(x)-cx^{-\alpha}|\le Cx^{-\alpha-\rho}$ for $x\ge x_0$. Then
\begin{equation}
\mathbb{E}[n_\nu]=\nu(\nu c)^{1/\alpha}\Gamma\!\left(1-\tfrac1\alpha\right)
\Bigl(1+O(\nu^{-\rho/\alpha})+O(\nu^{-1/\alpha})\Bigr),
\label{eq:remrate}
\end{equation}
and consequently
$\mathbb{E}[n_{s\nu}]/\mathbb{E}[n_\nu]=s^{1+\sigma}\bigl(1+O((s\nu)^{-\tau})\bigr)$
with $\tau=\min(\rho,1)/\alpha$, so $\delta_\nu=O((s\nu)^{-\tau})$.
\end{proposition}

The proof is in Appendix~\ref{app:proofs}; the step that matters is
pointwise bound $|e^{-a}-e^{-b}|\le|a-b|$ is too crude --- it leaves a term of order
$\nu$, which is larger than the main term for $\alpha>1$ --- and must be sharpened to
$|a-b|e^{-\min(a,b)}$, after which the substitution $t=\nu cx^{-\alpha}/2$ turns the
error integral into an incomplete Gamma function of order $1+(\rho-1)/\alpha$.

The rate is exact on our own family. For $\mu(x)=c(1+x)^{-\alpha}$ the second-order
condition holds with $\rho=1$, so $\tau=1/\alpha$; evaluating $\mathbb{E}[n_\nu]$ by
quadrature over $\nu\in[50,12800]$ gives measured exponents $-0.769$, $-0.625$ and
$-0.500$ at $\alpha=1.3,1.6,2.0$, against $-1/\alpha=-0.769,-0.625,-0.500$. With
\eqref{eq:remrate} every term of \eqref{eq:decomp3} is now accounted for: the
deterministic remainder at rate $(s\nu)^{-\min(\rho,1)/\alpha}$, the sampling
fluctuation at rate $(s\nu)^{1/2-1/\alpha}$, and the estimator error through
Proposition~\ref{prop:prop}. What remains outside the analysis is the behaviour of
$\hat\alpha$ itself, which we study empirically in Section~\ref{sec:limits}.

Second, the bound is loose and we say so rather than trading on the shape of its
condition. Sampling synthetic graphexes at five sizes and six tail indices, the
measured relative standard deviation of the edge count decays at $\nu^{-0.50}$ to
$\nu^{-0.60}$, matching \eqref{eq:rates}; that of the vertex count decays at
$\nu^{-0.65}$ to $\nu^{-0.75}$ across the whole range, including $\alpha=2.6$ where
\eqref{eq:rates} predicts no decay at all. The counts therefore concentrate well
beyond the regime the bound certifies, and the threshold $\alpha<2$ is an artefact of
the Poincar\'e step, not a phase boundary. What Proposition~\ref{prop:conc} buys is a
certificate rather than a rate: it converts the middle term of \eqref{eq:decomp3}
from an unexamined assumption into something that provably vanishes in the regime the
paper operates in, and it leaves $\delta_\nu$ as the single uncontrolled quantity in
the reduction.

\section{Method}
\label{sec:method}

\paragraph{Parameterisation.} With $u=\log(1+x)$ and a cut-off $U_R=\log(1+R)$, the
state is $\theta=(\lambda\in\mathbb{R}^{B},\alpha\in\mathbb{R},
\operatorname{vech}(\kappa))$, with $\lambda$ free on $B$ bulk cells and
$\lambda(u)=\lambda_B-\alpha(u-U_R)$ on the tail; $\kappa$ is free on the bulk and
zero outside. We use $B=16$ and $R=100$ (chosen by the ablation in
Section~\ref{sec:ablation}), giving $\dim\theta=153$. Flow matching runs on $\theta$
with the linear path of Remark~\ref{rem:cfm} and a four-layer MLP.

\paragraph{Estimation.} Training data are graphs, so each must be mapped to a
$\theta$. With $\nu=1$ (absorbing size into $\mu$), a node at $x$ has degree
$\mathrm{Poisson}(\mu(x))$ and the expected number of nodes of degree $k$ is
\begin{equation}
\label{eq:Nk}
N_k=\underbrace{\sum_{b}\mathrm{Poisson}(k;\mu_b)\Delta x_b}_{\text{bulk}}
+\underbrace{\frac{e^{U_R}\mu_R^{1/\alpha}}{\alpha\,k!}\,
\gamma\!\left(k-\tfrac1\alpha,\ \mu_R\right)}_{\text{tail}} ,
\end{equation}
the tail term following from $w=\mu_R e^{-\alpha v}$ and so available in
\emph{closed form}.

The degree tail this produces is a power law only on an intermediate range, and we
state the condition rather than an unqualified limit. For \emph{fixed} $\mu_R$ the
lower incomplete gamma $\gamma(k-1/\alpha,\mu_R)$ does not approach $\Gamma$ as
$k\to\infty$; instead $\gamma(a,x)\simeq x^{a}e^{-x}/a$, so
$N_k\propto\mu_R^{k}/k!$ and the decay is factorial, not polynomial. The power law
$N_k\propto k^{-1-1/\alpha}$ holds in the regime $1\ll k\lesssim\mu_R$, where
$\gamma\to\Gamma$, and is cut off beyond it --- which is exactly the finite-size
truncation of an observed degree tail. Numerically, the local log-log slope of
$N_k$ matches $-(1+1/\alpha)$ to within $0.15$ for $k\in[5,35]$ at $\mu_R=50$ and
across the whole tested range $k\in[5,400]$ at $\mu_R=500$, while at $\mu_R=5$ it
never does (slope $-9.3$ already at $k=10$). Since $\mu_R$ grows with the size
parameter, larger graphs widen the power-law window; the asymptotic statement is a
joint one in $(\nu,k)$, not a statement about $k$ alone.

\paragraph{What is exact here, and what is not.} Equation~\eqref{eq:Nk} gives
$\mathbb{E}[N_k]$ \emph{exactly}: it is an expectation over the Poisson process and
the conditionally independent edges, so linearity applies and no approximation
enters. The counts themselves, however, are \emph{not} independent across $k$, and
we do not claim otherwise. Degrees are not independent marks: the edge $(i,j)$
increments both $D_i$ and $D_j$, so the thinning theorem does not apply and
$\{N_k\}$ inherits a dependence through shared edges. Treating them as independent
Poisson therefore yields a \emph{composite} (pseudo-)likelihood
$\sum_k(N_k-n_k\log N_k)$, not an exact one. The dependence enters only through
individual edge indicators, each of probability $O(\psi_i\psi_j)$, which is the
regime in which composite likelihoods for sparse graphs are usually argued to be
well behaved; we use the objective as an estimating equation with the correct mean
structure and do not rely on its being a true likelihood anywhere in the paper. We
grid the scalar $\alpha$ (the incomplete gamma is not differentiable in its first
argument in standard libraries) and run batched Adam on $\lambda$ per grid value.

\paragraph{Anchored generation.} Generating at a target size requires a size
parameter $\nu$. Solving $n_\nu=n_{\mathrm{full}}$ directly from the fitted
parameters makes the edge count $e_\nu=\tfrac{\nu^2}{2}(\int\psi)^2$ inherit the
full scale error of the fit, and by Proposition~\ref{prop:prop} the effective
extrapolation ratio is then $n_{\mathrm{full}}$ itself rather than the ratio
$R=n_{\mathrm{full}}/n_s$ between the target and the training graph --- a factor
$\ln n_{\mathrm{full}}/\ln R \approx 3$--$7$ more sensitivity to the tail-index
error. We therefore \emph{anchor} generation to the training graph by imposing two
constraints instead of one: with $\psi\mapsto c\psi$,
\begin{align}
&\underbrace{\tfrac{\nu^2}{2}\Bigl(c\!\int\!\psi\Bigr)^{2} = e_s\,R^{\beta}}
   _{\text{edge count, from the observed } e_s} ,
\label{eq:anchor1}\\
&\underbrace{n_\nu[c\psi]=n_{\mathrm{full}}}_{\text{node count}} .
\label{eq:anchor2}
\end{align}
The first fixes the product $c\nu$, after which the second is a one-dimensional
root-find in $\nu$. Scaling $\psi$ alone does not suffice: re-solving $\nu$ from the
node-count equation cancels most of the rescaling, leaving a net dependence
$e\propto c^{\,2-4/(\alpha+1)}$ whose exponent is only $0.33$ at $\alpha=1.4$.

\paragraph{From $\mu$ to $\psi$: an approximation we make explicit.} The degree
likelihood constrains the marginal $\mu(x)=\int W(x,y)\dd y$, not $\psi$ itself. In
general $\mu(x)=\psi(x)\int\psi(y)K(x,y)\dd y$, so the two differ whenever $K\not\equiv1$.
We recover $\psi$ from the fitted $\mu$ under the rank-one relation
$\mu=\psi\int\psi$, giving $\int\psi=\sqrt{\int\mu}$ and $\psi=\mu/\!\int\!\psi$, and
we do \emph{not} re-impose marginal consistency after $\kappa$ is estimated. The
fitted object is therefore exactly the graphex we intend only when $K\equiv1$; with
a non-trivial bulk kernel the recovered $\psi$ absorbs part of $K$'s effect on the
marginal. We ablate this in Section~\ref{sec:fixedpoint}: iterating $\psi$ and $K$ to a
marginal-consistent fixed point is straightforward, and we report what it changes.

\paragraph{Estimating $\kappa$.} We invert the cumulative count
$C(u)=\int_0^u(1-e^{-\mu})e^{v}\dd v$ to assign positions, compare observed edge
counts between bulk cells against the rank-one prediction $\psi_i\psi_j$, and take
logs; estimation is on a coarse $4\times4$ grid, expanded block-constantly, with a
pseudo-count shrinking towards $\kappa=0$.

\section{Experiments}
\label{sec:exp}

\subsection{Path diagnostic}

\begin{figure}[t]
\centering
\includegraphics[width=0.8\columnwidth]{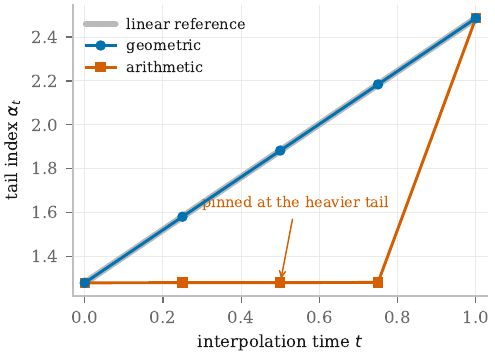}
\caption{Tail index along the interpolation path (Propositions~\ref{prop:arith}
and~\ref{prop:geom}). The geometric path coincides with the linear reference
exactly --- the grey band beneath it is that reference. The arithmetic path is
pinned at the heavier tail for every interior $t$ and jumps only at the endpoint,
so the quantity the model must learn is absent from the whole trajectory.}
\label{fig:path}
\end{figure}

Propositions~\ref{prop:arith} and~\ref{prop:geom} are directly observable.
Figure~\ref{fig:path} interpolates between two graphexes with $\alpha_0=1.278$ and
$\alpha_1=2.486$ and reports the tail index recovered by re-fitting the interpolant
at each $t$. The geometric path lands on the linear reference at every point; the
arithmetic path stays within $0.002$ of the heavier tail across the whole interior
and reaches $\alpha_1$ only at the endpoint.

\subsection{Verifying the two analytic predictions}
\label{sec:verify}

\paragraph{Clustering identity.} Proposition~\ref{prop:trans} predicts a
transitivity independent of $\nu$. Sampling rank-one graphexes at three sizes gives
a predicted value of $0.02897$ at every size, against measured $0.0326$, $0.0304$
and $0.0301$ at $n=300$, $1000$ and $3000$ --- ratios $1.13$, $1.05$, $1.04$. Both
the value and its independence of $\nu$ are confirmed.

\begin{figure}[t]
\centering
\includegraphics[width=0.8\columnwidth]{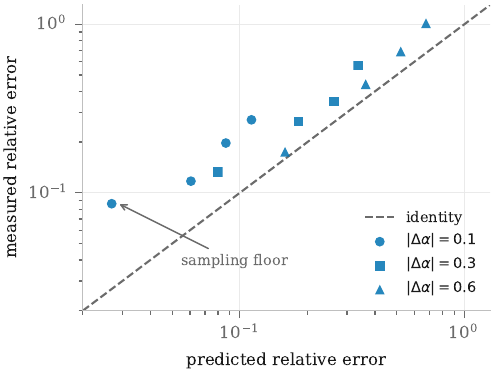}
\caption{Verification of Proposition~\ref{prop:prop}. Measured against predicted
relative edge error over extrapolation ratios $2$--$25\times$ and tail-index errors
$0.1$--$0.6$ ($r=0.978$). Points approach the identity as the first-order term
grows; the systematic excess at small $|\Delta\alpha|$ is the sampling floor, which
does not vanish with the tail-index error.}
\label{fig:prop}
\end{figure}

\paragraph{Error propagation.} Sweeping the tail-index error and the extrapolation
ratio, the measured relative edge error correlates with the prediction of
Proposition~\ref{prop:prop} at $r=0.978$. The residual is structured as expected:
the ratio of measured to predicted error falls from $3.2$ at
$|\Delta\alpha|=0.1$ to $1.10$ at $|\Delta\alpha|=0.6$, i.e.\ the first-order term
governs once it exceeds the sampling floor.

\subsection{Synthetic round trip}

We synthesise $5000$ graphexes, sample a graph from each at $n\approx230$, estimate,
train the flow, generate, and sample at $n\approx2518$: a $10.9\times$
extrapolation with analytic ground truth throughout. Three seeds, $600$ evaluation
samples per arm.

\begin{table}[t]
\caption{Synthetic round trip at $10.9\times$. \emph{oracle params} isolates the
flow by training on true parameters; \emph{full pipeline} estimates everything from
observed graphs. The graphon edge error is an identity, not a fitted quantity.}
\label{tab:round}
\centering
\resizebox{\columnwidth}{!}{%
\begin{tabular}{lccc}
\toprule
& $\beta$ & $\alpha$ & invalid \\
\midrule
ground truth                  & $1.270$ & $1.919\pm0.384$ & --- \\
\midrule
geometric, oracle params      & $\mathbf{1.276}\pm0.002$ & $\mathbf{1.922}\pm0.384$ & $0\%$ \\
arithmetic, oracle params     & $1.282\pm0.001$ & $3.056\pm1.150$ & $\mathbf{27\%}$ \\
full pipeline (all estimated) & $1.283\pm0.003$ & $1.835\pm0.304$ & $0\%$ \\
\midrule
graphon baseline              & $2$ (identity) & --- & edge error $473\%$ \\
\bottomrule
\end{tabular}}
\end{table}

With oracle parameters the flow reproduces not only the mean exponent but its
dispersion ($0.384$ against a true $0.384$), so the flow is not the bottleneck. The
arithmetic path fails specifically: a quarter of generated objects are invalid, the
exponent is over-estimated by $59\%$, and its dispersion is inflated threefold.

\paragraph{An evaluation-protocol warning.} The arithmetic arm's $\beta$ is within
$1\%$ of truth; reported alone it would look adequate. Conversely
Table~\ref{tab:struct} shows it matching or beating the geometric arm on
\emph{fixed-size} structural statistics --- because $\alpha$ governs scaling with
$\nu$, not shape at one size, and because those statistics are computed only over
the $73\%$ of samples that are valid. Either metric alone supports a wrong
conclusion, so we report scaling, structure and validity together.

\begin{table}[t]
\caption{Structural statistics on the large graphs. degW1 is the $1$-Wasserstein
distance between degree distributions on $\log(1+d)$. The arithmetic arm is
competitive here while being badly wrong in Table~\ref{tab:round}.}
\label{tab:struct}
\centering
\resizebox{\columnwidth}{!}{%
\begin{tabular}{lccccc}
\toprule
& transitivity & assortativity & LCC frac. & mean deg. & degW1 \\
\midrule
ground truth            & $0.0366$ & $0.0712$ & $0.848$ & $5.00$ & --- \\
geometric, oracle       & $0.0363$ & $0.0771$ & $0.832$ & $4.77$ & $0.048$ \\
arithmetic, oracle      & $0.0353$ & $0.0886$ & $0.841$ & $4.53$ & $0.029$ \\
full pipeline           & $0.0256$ & $-0.0109$ & $0.918$ & $5.00$ & $0.021$ \\
\bottomrule
\end{tabular}}
\end{table}

\subsection{Is the flow doing anything a simpler density could not?}
\label{sec:thetabase}

Showing that arithmetic interpolation fails is not the same as showing that flow
matching is needed: $\theta$ is a $153$-dimensional vector, and a Gaussian or a kernel
density over it would be far cheaper. We therefore fit four alternatives to the same
synthetic $\theta$ population and compare them on the four things the construction is
supposed to deliver.

\begin{table}[t]
\caption{Generators over $\theta$-space, trained on $2250$ synthetic graphexes and
evaluated against $750$ held out. \emph{valid} is the fraction of generated $\theta$
that are admissible graphexes; \emph{energy} is the energy distance to the held-out
set, whose achievable floor --- the training sample against the held-out sample ---
is $0.0319$. \emph{bootstrap} resamples training $\theta$ and is a memorisation
reference, not a usable model.}
\label{tab:thetabase}
\centering
\resizebox{\columnwidth}{!}{%
\begin{tabular}{lccccc}
\toprule
generator & valid & $\alpha$ mean & $\alpha$ sd & energy & $\beta$ \\
\midrule
ground truth        & ---            & $1.924$ & $0.400$ & $0.0319$ & $1.265$ \\
\midrule
flow (ours)         & $\mathbf{100\%}$ & $1.913$ & $0.390$ & $\mathbf{0.0372}$ & $1.272$ \\
full-covariance Gaussian & $98\%$    & $1.923$ & $0.375$ & $0.0761$ & $1.294$ \\
diagonal Gaussian   & $0\%$          & $1.938$ & $0.380$ & $0.3868$ & $1.265$ \\
kernel density      & $0\%$          & $1.989$ & $0.500$ & $0.4188$ & $1.268$ \\
\midrule
bootstrap \emph{(memorisation)} & $100\%$ & $1.926$ & $0.392$ & $0.0503$ & $1.271$ \\
\bottomrule
\end{tabular}}
\end{table}

Two things separate the flow, and neither is the exponent. Every generator reproduces
the mean tail index and most reproduce its dispersion, so a comparison restricted to
$\alpha$ or $\beta$ --- which is what the rest of this paper measures --- would
conclude that the choice of generator does not matter.

\emph{Validity is not automatic.} A diagonal Gaussian and a kernel density produce
\emph{no} admissible graphex at all. The constraint set is not a product: $\lambda$
must be non-increasing across $16$ bulk cells and $\ell=\lambda_i+\lambda_j+\kappa_{ij}
\le0$ must hold on all pairs, so coordinate-independent noise violates something
almost surely. A full-covariance Gaussian recovers most of this ($98\%$) because the
constraints are close to linear, and the flow recovers all of it. This is the concrete
sense in which the generative half is not decoration: the object being generated lives
on a constrained set, and the two cheapest densities never land on it.

\emph{The flow is closer to the target than resampling the data.} Its energy distance
to the held-out population, $0.0372$, is nearer the achievable floor of $0.0319$ than
the bootstrap's $0.0503$, and half the full-covariance Gaussian's $0.0761$. Beating
the bootstrap is the informative comparison, since bootstrap is what one gets by
memorising the training set: the flow is interpolating the population rather than
reproducing the sample.

We state the scope of this result exactly: it is synthetic, where ground truth is
analytic, and it compares generators of $\theta$, not of graphs. It establishes that
the flow earns its place over the obvious cheaper densities \emph{on this
population}; it does not establish a benefit on real corpora, which remains open
(Section~\ref{sec:limits}).

\subsection{Bulk capacity ablation}
\label{sec:ablation}

Under the $\nu=1$ convention the number of observable bulk nodes is bounded by
$\int_0^{U_R}e^{u}\dd u=R$, independently of graph size, so $R$ is a genuine
capacity parameter. Table~\ref{tab:R} sweeps it.

\begin{table}[t]
\caption{Bulk capacity $R$. Scaling accuracy is $U$-shaped with an optimum at
$R=100$; structure improves only marginally and assortativity never recovers its
sign, consistent with Proposition~\ref{prop:trans}.}
\label{tab:R}
\centering
\resizebox{\columnwidth}{!}{%
\begin{tabular}{lccccc}
\toprule
$R$ & bulk nodes & $\beta$ MAE & transitivity & assortativity & degW1 \\
\midrule
$25$  & $24$  & $0.100$ & $0.032$ & $-0.023$ & $0.046$ \\
$50$  & $47$  & $0.064$ & $0.026$ & $-0.020$ & $0.021$ \\
$100$ & $86$  & $\mathbf{0.045}$ & $0.025$ & $-0.011$ & $0.046$ \\
$200$ & $132$ & $0.084$ & $0.026$ & $-0.005$ & $0.040$ \\
$400$ & $173$ & $0.157$ & $0.024$ & $-0.006$ & $0.069$ \\
\midrule
\multicolumn{2}{l}{ground truth} & --- & $0.037$ & $0.070$ & --- \\
\bottomrule
\end{tabular}}
\end{table}

Increasing capacity sevenfold moves assortativity from $-0.023$ to $-0.006$ without
changing its sign and leaves transitivity essentially flat. Capacity is therefore
\emph{not} the binding constraint on structure; Proposition~\ref{prop:trans} is.
What $R$ does buy is scaling accuracy, and we use $R=100$ throughout.

\subsection{Does the rank-one inversion matter?}
\label{sec:fixedpoint}

Section~\ref{sec:method} recovers $\psi$ from the fitted marginal under the
rank-one relation, which is inexact once $K\not\equiv1$. Iterating
$\psi^{t+1}(x)=\hat\mu(x)\big/\!\int\!\psi^{t}(y)K(x,y)\dd y$ to a marginal-consistent
fixed point and re-estimating $K$ costs almost nothing; Table~\ref{tab:fixedpoint}
reports what it buys.

\begin{table}[t]
\caption{Single-pass rank-one inversion against iterating $\psi$ and $K$ to
marginal consistency. The tail index is untouched --- it comes from the degree fit,
not the inversion --- while one iteration improves structural fidelity.}
\label{tab:fixedpoint}
\centering
\footnotesize
\resizebox{\columnwidth}{!}{%
\begin{tabular}{lccccc}
\toprule
& $\beta$ MAE & transitivity & assortativity & mean deg. & degW1 \\
\midrule
ground truth        & ---      & $0.0346$ & $0.0687$  & $4.68$ & --- \\
\midrule
single-pass         & $0.0500$ & $0.0236$ & $-0.0101$ & $4.77$ & $0.0331$ \\
fixed point, $1$ it.& $0.0500$ & $\mathbf{0.0300}$ & $-0.0097$ & $5.01$ & $\mathbf{0.0242}$ \\
fixed point, $3$ it.& $0.0500$ & $0.0307$ & $-0.0105$ & $5.08$ & $0.0322$ \\
\bottomrule
\end{tabular}}
\end{table}

The tail index is unchanged to four decimals, which is the expected behaviour: it is
identified by the degree histogram and never passes through the inversion. One
iteration moves transitivity from $32\%$ below truth to $13\%$ below and cuts the
degree-distribution distance by $27\%$; further iterations do not help. The
assortativity error is untouched, consistent with the reading in
Section~\ref{sec:gen} that it is not a marginal-consistency problem. All tables
elsewhere in the paper use the single-pass version, so the structural numbers we
report are conservative by roughly this margin; the headline extrapolation results
are unaffected either way.

\subsection{The design-based view: what our estimator is}
\label{sec:design}

It is worth stating plainly what the estimator of Section~\ref{sec:method} is,
because in one reading it is not a new estimator at all.

Under uniform node $p$-sampling with retention probability $s$, an edge survives
exactly when both of its endpoints do, with probability $s^2$. Hence
$\mathbb{E}[e_s]=s^2e_1$ and the design-based estimator
\begin{equation}
\hat e^{\,\mathrm{des}} = e_s/s^{2}
\label{eq:design}
\end{equation}
is unbiased and requires no model whatsoever. If $s$ is known, this is the estimator
to use --- it reaches a median error of $7.1\%$ on our benchmark against $26.8\%$
for the best inferred exponent --- and we report it throughout as an oracle.

Our setting is that $s$ is \emph{not} known while the population size $n_1$ is.
This is the common situation in the applications that motivate the paper: a released
partial dump, a panel exposed through an API, a crawl stopped by a budget --- in
none of these does the analyst set the inclusion probability, while $n_1$ is often
known from an external register (allocated AS numbers, registered accounts, indexed
documents). The question is then whether $s$ can be recovered from what is observed.

\paragraph{Which $n_1$, exactly.} The quantity the theory uses is the number of
\emph{non-isolated} vertices of the target, since $n_\nu$ counts exactly those and
sub-sampling discards the vertices it leaves isolated. An external register need not
supply that number: a platform's registered-account count includes accounts with no
interactions, and an index's document count includes documents with no links. Where
the two differ, using the register in place of the non-isolated count inflates $n_1$,
which inflates $R$ and, through \eqref{eq:shat}, deflates $\hat s$; both push the
edge estimate up. The requirement is therefore sharper than ``the population size is
known'': it is that the size of the interacting population is known, and an analyst
who has only a register should treat the difference as a bias in $R$ of the same form
as an error in $\hat\sigma$, with the sensitivity given by
Proposition~\ref{prop:prop}. Our benchmark satisfies the requirement by construction
--- the SNAP graphs are given as edge lists, so every vertex is non-isolated --- which
means our experiments do not test this distinction, and we do not claim they do.

It can, asymptotically, and this is where the tail index enters. By
Proposition~\ref{prop:scaling} the expected non-isolated vertex count satisfies
$\mathbb{E}[n_\nu]=C\nu^{1+\sigma}(1+o(1))$ with $\sigma=1/\alpha$, so sub-sampling at
rate $s$ gives $\mathbb{E}[n_{s\nu}]/\mathbb{E}[n_\nu]\to s^{1+\sigma}$. Treating this
as an identity on the observed graph gives
\begin{equation}
\hat s = \left(n_s/n_1\right)^{1/(1+\sigma)} .
\label{eq:shat}
\end{equation}
Substituting \eqref{eq:shat} into \eqref{eq:design} gives
$e_s(n_1/n_s)^{2/(1+\sigma)}=e_s R^{\beta}$: our estimator and the design-based
estimator are the same expression, differing only in whether $s$ is supplied or
inferred. We verified the substitution numerically as a check on the implementation;
it agrees to machine precision ($4.5\times10^{-16}$ relative).

Two things in that sentence should not be conflated. The \emph{substitution} is
exact algebra. The \emph{relation} \eqref{eq:shat} is not: it is the finite-$\nu$ use
of a limit statement about expectations, so $\hat s$ inherits both an approximation
error, which we do not bound, and the error in $\hat\sigma$, which we do measure. The
two together are what Table~\ref{tab:baselines} reports, and separating them is open.
Empirically the combined effect is a median $\hat s/s$ of $1.136$ over the $39$
settings --- an over-estimate of $14\%$ --- against $0.532$ for the naive plug-in
below.

Three consequences follow, and they organise the rest of the paper.

\emph{The graphon estimator is the naive plug-in.} Setting $\sigma=0$ in
\eqref{eq:shat} gives $\hat s=n_s/n_1$ --- the observed vertex fraction, used
directly. That is exactly the $\beta=2$ estimator. Its failure is therefore not a
fitting failure but the omission of the isolate correction: sub-sampling leaves
vertices isolated and they leave the observable graph, so $n_s/n_1$ systematically
underestimates $s$. Measured over the $39$ settings, $n_s/n_1$ recovers a median of
$0.532$ of the true $s$, while \eqref{eq:shat} recovers $1.136$ --- an error of
$47\%$ against one of $14\%$, which squares into the gap between $260\%$ and
$27\%$ in edge count.

\emph{Estimating the exponent is estimating the sampling rate.} Everything the
paper does to the tail index is, in this view, an attempt to recover the inclusion
probability of an observation process the analyst did not design. This is why the
comparison that matters is against other tail-index estimators
(Section~\ref{sec:baselines}) rather than against graph generative models.

\emph{The oracle bounds the family.} Because \eqref{eq:design} with the true $s$ is
the best any member of this family can do, the gap between it and our estimator is
exactly the price of not knowing $s$, and we report it as such.

\subsection{Real networks: out-of-sample extrapolation}
\label{sec:real}

The synthetic round trip validates internal consistency; it cannot establish that
real networks lie in this class. Using the sub-sampling equivalence of
Section~\ref{sec:bg}, we retain each node with probability $s$, fit the tail index
\emph{only on the sub-sample}, and predict the full-graph edge count as
$e_s(n_1/n_s)^{\beta}$. The graphon baseline is the same expression with exponent
$2$.

\begin{figure}[t]
\centering
\includegraphics[width=0.8\columnwidth]{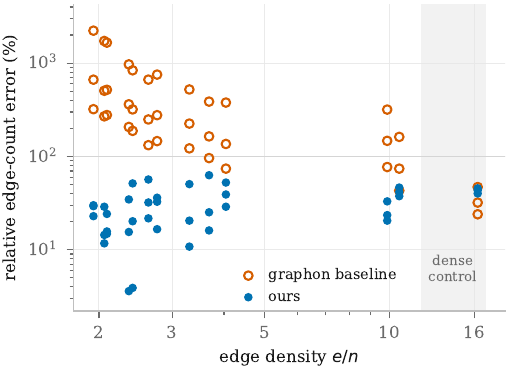}
\caption{Out-of-sample extrapolation error across $13$ real networks and $39$
fit/extrapolate settings, against edge density. The method is ahead by roughly an
order of magnitude throughout the sparse regime, the two curves meet near
$e/n\approx10$, and inside the dense control ($e/n=16.3$) the graphon baseline is
the better model --- the crossover the construction predicts.}
\label{fig:real}
\end{figure}

\begin{table*}[t]
\caption{Out-of-sample extrapolation on real networks: fit the tail index on a
sub-sample only, predict the full-size edge count. All numbers are means over the $15$ sub-sampling
seeds of Table~\ref{tab:baselines}, so they are directly comparable to the baselines
there. \texttt{email-Eu-core} is a deliberate \emph{dense} control. All $13$ networks
are listed, ordered by edge density.}
\label{tab:real}
\centering
\footnotesize
\begin{tabular}{lcccc}
\toprule
dataset ($n$) & $e/n$ & extrapolation & ours & graphon \\
\midrule
\texttt{as20000102} ($6.5k$) & $1.94$ & $53/17/8\times$ & $18.9/22.1/23.3\%$ & $2079.3/659.8/311.4\%$ \\
\texttt{oregon1\_010331} ($11k$) & $2.06$ & $57/10/6\times$ & $29.9/24.5/21.2\%$ & $2000.0/560.9/277.9\%$ \\
\texttt{oregon1\_010526} ($11k$) & $2.09$ & $57/10/6\times$ & $21.6/20.1/21.5\%$ & $2147.9/629.9/312.8\%$ \\
\texttt{p2p-Gnutella31} ($63k$) & $2.36$ & $34/11/6\times$ & $45.8/16.9/3.7\%$ & $954.4/365.0/210.4\%$ \\
\texttt{p2p-Gnutella25} ($23k$) & $2.41$ & $30/10/6\times$ & $44.6/21.0/8.4\%$ & $860.3/325.7/186.8\%$ \\
\texttt{ca-HepTh} ($9.9k$) & $2.63$ & $27/9/5\times$ & $36.7/44.3/23.1\%$ & $760.3/255.6/132.4\%$ \\
\texttt{ca-GrQc} ($5.2k$) & $2.76$ & $26/9/5\times$ & $22.2/36.8/26.8\%$ & $767.6/260.3/136.7\%$ \\
\texttt{p2p-Gnutella08} ($6.3k$) & $3.30$ & $24/8/5\times$ & $31.7/41.7/12.6\%$ & $563.4/214.3/121.1\%$ \\
\texttt{p2p-Gnutella04} ($11k$) & $3.68$ & $22/8/5\times$ & $54.4/32.0/15.1\%$ & $412.9/157.4/94.9\%$ \\
\texttt{ca-CondMat} ($23k$) & $4.04$ & $21/8/4\times$ & $62.1/39.9/29.6\%$ & $399.5/138.4/71.6\%$ \\
\texttt{ca-HepPh} ($12k$) & $9.87$ & $19/7/4\times$ & $44.3/23.7/21.0\%$ & $318.0/137.8/76.4\%$ \\
\texttt{ca-AstroPh} ($19k$) & $10.6$ & $15/6/4\times$ & $43.6/39.1/35.8\%$ & $166.5/73.8/42.1\%$ \\
\texttt{email-Eu-core} ($1.0k$) & $16.3$ & $14/7/4\times$ & $66.2/28.7/21.3\%$ & $78.9/41.5/31.9\%$ \\
\midrule
\multicolumn{3}{l}{median over all $39$ settings ($13$ networks $\times$ $3$ fractions)} & $\mathbf{26.8\%}$ & $260\%$ \\
\bottomrule
\end{tabular}
\end{table*}

The method is ahead of the graphon in $36$ of $39$ settings; the exceptions are
the dense control. The graphon is not the demanding comparison, however, and the
numbers here should be read together with Section~\ref{sec:baselines}, which
replaces it with a direct log--log regression ($35.0\%$), the published graphex
tail-index estimator ($21.7$--$24.8\%$, better than ours) and an oracle that knows
the sampling rate ($7.1\%$).
The crossover is near $e/n\approx12$--$16$: at $e/n\approx10$
(\texttt{ca-HepPh}, \texttt{ca-AstroPh}) the method is still ahead, and at $16.3$ it
is not --- where the graphon baseline is the correct model. The fitted tail index
detects the regime, returning $\alpha\in[2.4,4.2]$ for the dense networks against
$[1.3,1.7]$ for the sparse ones. On the largest network tested ($62{,}586$ nodes)
the error at a $5.9\times$ extrapolation is $3.7\%$.

\subsection{How much of this needs a model?}
\label{sec:baselines}

The graphon baseline fails by an identity, so beating it establishes only that a
fixed graphon is the wrong object for sparse extrapolation. The question that
decides whether the model earns its cost is what \emph{else} produces the exponent,
including methods that use no model at all. We compare seven ways of obtaining
$\beta$ in the same prediction $e_s(n_1/n_s)^{\beta}$, so that only its provenance
differs, plus the design-based oracle of Section~\ref{sec:design}, which is not of
that form.

\begin{table*}[t]
\caption{Where the accuracy comes from, over $13$ networks $\times$ $3$ budgets and
$15$ sub-sampling seeds per setting. All rows but the last use the same prediction
$e_s(n_1/n_s)^{\beta}$ and differ only in how $\beta$ is obtained; the last knows the
true inclusion probability. Median and mean are over the $39$ settings. The test is
\emph{network-level}: errors are averaged within each network first, since the three
budgets share a graph and are not independent, and a two-sided Wilcoxon signed-rank
test is then run on the $13$ paired values. ``wins'' counts networks on which ours is
ahead. The last two columns split the settings at $e/n=3$.}
\label{tab:baselines}
\centering
\footnotesize
\begin{tabular}{llcccccc}
\toprule
$\beta$ from & & median & mean & $p$ (net-level) & wins & sparse & dense \\
\midrule
constant mean degree & $\beta=1$ & $62.2\%$ & $61.9\%$ & $2.4\!\times\!10^{-4}$ & $13/13$ & $55.0\%$ & $69.2\%$ \\
Hill tail-index estimator & & $61.4\%$ & $61.0\%$ & $2.4\!\times\!10^{-4}$ & $13/13$ & $53.5\%$ & $68.5\%$ \\
fixed graphon & $\beta=2$ & $260.3\%$ & $444.5\%$ & $2.4\!\times\!10^{-4}$ & $13/13$ & $365.0\%$ & $129.4\%$ \\
log--log regression on internal sub-samples & & $35.0\%$ & $38.9\%$ & $1.7\!\times\!10^{-3}$ & $12/13$ & $33.0\%$ & $43.8\%$ \\
\midrule
degree-distribution fit (ours) & & $26.8\%$ & $30.2\%$ & --- & --- & $\mathbf{22.2\%}$ & $33.9\%$ \\
graphex tail index \cite{naulet2021bootstrap}, $p=0.8$ & & $24.8\%$ & $26.9\%$ & $9.4\!\times\!10^{-2}$ & $3/13$ & $22.9\%$ & $24.9\%$ \\
graphex tail index \cite{naulet2021bootstrap}, $p\to1$ & & $\mathbf{21.7\%}$ & $\mathbf{24.6\%}$ & $6.1\!\times\!10^{-3}$ & $2/13$ & $\mathbf{21.4\%}$ & $\mathbf{21.9\%}$ \\
\midrule
\emph{oracle}: $e_s/s^2$ with $s$ known & & $7.1\%$ & $13.0\%$ & $3.4\!\times\!10^{-3}$ & $3/13$ & $7.1\%$ & $6.1\%$ \\
\bottomrule
\end{tabular}
\end{table*}

Four readings, and two of them go against us.

\emph{Most of the distance is bought by any exponent below two.} Measuring the
densification exponent by direct log--log regression already reaches $35.0\%$ from
the graphon's $260\%$. Whatever else is true, the substantive claim of this paper is
the reduction, not any particular estimator of $\beta$.

\emph{Our estimator is not the best one for edge count.} The published graphex
tail-index estimator of \cite{naulet2021bootstrap} reaches $24.8\%$ at $p=0.8$ and
$21.7\%$ in its $p\to1$ form, against our $26.8\%$. Tested at network level the
first gap is \emph{not} significant ($p=0.094$, ours ahead on $3$ of $13$ networks)
while the second is ($p=6.1\times10^{-3}$, ours ahead on $2$); an earlier
setting-level test made the first look significant, which it is not once the three
budgets sharing a graph are pooled. We report the comparison rather than omit it. Two things qualify it. Restricted to the $21$
settings with $e/n<3$ --- the sparse regime the model is for --- the two are within
$0.7$ percentage points of each other ($22.2\%$ against $22.9\%$); the deficit is
concentrated in the denser networks, where our fitted $\alpha$ saturates (median
$1.62$ against $1.84$ from the tail-index estimator, correlation $0.92$), so
$\beta=2\alpha/(\alpha+1)$ comes out too small and the edge count is under-predicted.
And the two estimators are not interchangeable in what they return: ours yields a
degree distribution at any size and a generative object, which
Section~\ref{sec:degdist} shows no exponent-only method can supply. A practitioner
who wants only an edge count should use \cite{naulet2021bootstrap}; the fit is worth
its cost when more than the count is needed.

\emph{The $p\to1$ limit of that estimator is the fraction of degree-one vertices.}
Writing $\varepsilon=1-p$ and expanding
$N_p=p\sum_j\mathsf{d}_j(1-(1-p)^j)$ gives
$\hat\sigma_p\to \mathsf{d}_1/n_s$. So the strongest method in the table is, in
closed form, ``take the proportion of degree-one vertices as $\sigma$ and predict
$e_s(n_1/n_s)^{2/(1+\sigma)}$'' --- $21.7\%$ median, no fitting of any kind. We
verified the limit numerically (agreement to four decimals at $p=0.999$) and we
state it because it is the most useful single sentence in this section for a
practitioner.

\emph{Knowing $s$ is worth more than any modelling.} The oracle reaches $7.1\%$,
three times better than the best inferred exponent. This is the honest measure of
what the tail index is being asked to do: recover a sampling rate. Its variance is
not negligible either --- the mean error is $13.0\%$ against a median of $7.1\%$,
because in a hub-dominated graph the loss of one high-degree vertex removes many
edges at once --- but nothing in this paper closes that gap. If the sample was
designed, use \eqref{eq:design}.

Two baselines that might have been expected to do well do not. Assuming a constant
mean degree ($\beta=1$) is never competitive, because these networks genuinely
densify; and the classical Hill estimator is no better, for the reason recorded in
Section~\ref{sec:limits} --- at these graph sizes the degree range spans too few
decades for an estimator that uses only the extreme order statistics. It is worth
noting that the two estimators that do work both use the \emph{bulk} of the degree
distribution, not its extreme tail.

\paragraph{On finite-sample guarantees.} Proposition~\ref{prop:scaling} is an
expectation-level statement and we prove no concentration; the estimator is applied
to a single graph, and its objective is a composite likelihood for dependent degree
counts, so neither the exponent nor the resulting edge count comes with a
finite-sample guarantee here. What we can offer is measurement. The median error
falls monotonically with the sampling budget (Section~\ref{sec:budget}), and the
measured error tracks Proposition~\ref{prop:prop}'s prediction at $r=0.978$ across
extrapolation ratios and tail-index errors. We deliberately do not report a
seed-to-seed variance for $\hat\alpha$ as evidence of a rate: the tail index is
selected on a $30$-point grid, so its spread across seeds is quantised by the grid
rather than by the data, and reading a convergence rate off it would be an artefact.

\subsection{The reduction is modular, and where the modularity stops}
\label{sec:modular}

Nothing in Section~\ref{sec:design} requires the tail index to come from our fit.
\eqref{eq:shat} takes a $\hat\sigma$ and returns an estimator; how $\hat\sigma$ was
obtained is a separate question, and the results above are best read as a comparison
of what to put in that slot rather than as a contest our estimator lost. Filled with
the closed-form estimator of \cite{naulet2021bootstrap} in its $p\to1$ form, the
reduction gives a median edge error of $21.7\%$ and needs no fitting of any kind:
count the degree-one vertices, divide by $n_s$, and evaluate
$e_s(n_1/n_s)^{2/(1+\hat\sigma)}$. That is the configuration we recommend for edge
count, and it is a result about the reduction, not about us.

The modularity has a limit, and it is worth reporting because it was not what we
expected. Substituting the same $\hat\sigma$ into the \emph{degree-profile} fit ---
holding $\alpha=1/\hat\sigma$ fixed and fitting only the bulk --- makes the estimated
degree distribution worse, not better: median $1$-Wasserstein $0.66$ against $0.62$
for our own jointly fitted index, and worse on six of the seven networks. The two
global statistics want different tail indices. The one that best reproduces how the
vertex count scales with $\nu$ is not the one that best fits the degree histogram at
a single size, because the first is a statement about the extreme tail and the second
is dominated by the bulk. A single object serving both must trade them off, which is
what our fit does and why it is not simply dominated by the specialised estimators.

\subsection{When the sample is a crawl, not a $p$-sample}
\label{sec:crawl}

Everything above assumes the observed graph is a uniform $p$-sample. Real crawls are
not: breadth-first and random-walk exploration reach high-degree vertices first, and
the induced subgraph is not distributed as a graphex sub-sample of the target. Since
crawls appear in the motivation, we measured the gap rather than arguing it away,
repeating the extrapolation on six networks with snowball and random-walk sampling at
the same vertex budgets.

Everything fails, so we report the conclusion rather than a table of unusable
numbers. Median error rises from $31.4\%$ under uniform sampling to $176.6\%$ under
snowball and $77.0\%$ under a random walk; the same comparison for the tail-index
estimator of \cite{naulet2021bootstrap} is $26.9\%\to324.7\%\to183.0\%$, and for the
design-based oracle $7.8\%\to588.1\%\to341.5\%$. The mechanism is visible in the
samples: a snowball subgraph is $1.36$ times denser than the graph it came from
against $0.40$ for a uniform sample of the same budget, and the fitted tail index
rises from $1.58$ to $2.48$ --- the crawl manufactures the appearance of a lighter
tail, the effect \cite{stumpf2005subnets} and \cite{achlioptas2009bias} document for
scale-free degree estimates and traceroute exploration. The estimator is not
malfunctioning; it is correctly fitting a graph that is not a graphex sub-sample.

The oracle degrading worst is the part worth keeping. It is the sharpest statement
available about the design-based estimator of Section~\ref{sec:design}: $e_s/s^2$ is
unbiased only because $s^2$ is the edge inclusion probability under uniform
independent retention, and under a crawl the fraction of vertices reached carries
almost no information about the fraction of edges seen. Knowing how much of the graph
you touched is worth a great deal when you designed the touching and very little when
you did not. The scope of this paper is therefore uniform $p$-sampling, which the
abstract and Section~\ref{sec:bg} state; pairing a graphex fit with a design
correction for degree-biased exploration --- the Horvitz--Thompson-style corrections
of \cite{ribeiro2010estimating, gjoka2010walking} are the natural ingredient --- is
left open.

\subsection{Sampling budget and estimation error}
\label{sec:budget}

\begin{figure}[t]
\centering
\includegraphics[width=0.8\columnwidth]{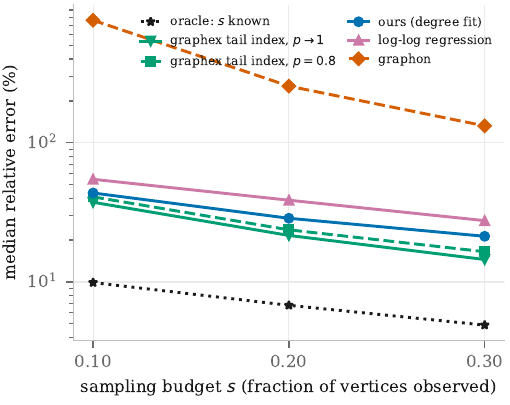}
\caption{Sampling budget against estimation error, median over $13$ networks at each
budget. The ordering of methods is unchanged across the range, so an estimator can be
chosen without knowing the budget in advance. The gap to the oracle --- the price of
not knowing the sampling rate --- is roughly a factor of four and does not close as
the budget grows.}
\label{fig:budget}
\end{figure}

The quantity a practitioner controls is the sampling budget: what fraction $s$ of the
vertices the crawl reaches before it stops. Figure~\ref{fig:budget} reads the $39$
settings as a budget--error curve, reporting the median relative error in estimated
global edge count at each budget.

Three features are worth naming. The ordering is stable across budgets, so choosing
an estimator does not require knowing the budget in advance. The three tail-index
methods stay within $6$--$7$ percentage points of each other throughout ($43.6\%$,
$40.9\%$ and $37.4\%$ at $s=0.10$; $21.3\%$, $16.5\%$ and $14.5\%$ at $s=0.30$),
which is the regime Proposition~\ref{prop:prop} describes --- they differ only
through $\Delta\beta$, and the resulting error scales as $R^{\Delta\beta}$ --- while
the separation from the log--log regression and the graphon is an order of magnitude
larger. The choice among tail-index estimators is a refinement; the choice to use one
at all is not. And the graphon's steep improvement is not evidence that it becomes
competitive: its error is $R^{2-\beta}$ by the identity of Section~\ref{sec:design},
so it necessarily improves as the budget grows and the extrapolation shortens, while
remaining wrong by construction.

The gap to the oracle is the more informative reading. It stays near a factor of
four at every budget ($9.9\%$ against $37.4\%$ at $s=0.10$, $4.9\%$ against $14.5\%$
at $s=0.30$): a larger budget makes the inference problem easier but does not make
the sampling rate observable, and no amount of budget substitutes for knowing it.

\subsection{Estimating the degree distribution}
\label{sec:degdist}

Edge count is one global statistic; the degree distribution is the other one a
sampled-graph analysis usually wants, and it is the task on which a fitted graphex
might earn its cost, since an exponent alone does not produce one. Four alternatives
have to be cleared. Two use no model: the observed sample's degree distribution as it
stands, and a thinning correction, since a retained vertex of degree $d$ has sampled
degree $\mathrm{Binomial}(d,s)$ so that $d_s/s$ is unbiased for $d$ when $s$ is known.
The third is the graphon. The fourth is the closest prior method: \cite{zhang2015estimating}
pose degree-distribution recovery from a sampled network as an inverse problem
$\mathbb{E}[n]=AN$, with $A_{kd}=s\binom{d}{k}s^k(1-s)^{d-k}$ under our design, and
solve it by non-negative penalised weighted least squares with the penalty parameter
chosen by Monte Carlo SURE. We implemented it, with two adaptations we state
because they both help it: the support is a $40$-point geometric grid in $d$ rather
than every integer, since a second-difference penalty on a linear grid is a poor prior
for a heavy tail and drives the solution to a smooth ramp; and we supply the known
population size as a constraint $\sum_d N_d=n_1$, which our estimator also uses. We
run it twice, once with the true $s$ and once with the $\hat s$ of \eqref{eq:shat},
the latter being the comparison at equal information.

\begin{table}[t]
\caption{Degree-distribution estimation at full size from a $10\%$ sample,
$1$-Wasserstein distance on $\log(1+d)$, means over $15$ seeds. Lower is better; bold
marks the best method that does not know $s$. The two right-hand columns are oracles.}
\label{tab:degdist}
\centering
\resizebox{\columnwidth}{!}{%
\begin{tabular}{lcccc|cc}
\toprule
 & \multicolumn{4}{c|}{$s$ unknown} & \multicolumn{2}{c}{$s$ known} \\
network & ours & \cite{zhang2015estimating} & sample & graphon
        & \cite{zhang2015estimating} & $d_s/s$ \\
\midrule
\texttt{as20000102}     & $\mathbf{0.17}$ & $0.35$ & $0.30$ & $2.83$ & $0.37$ & $1.47$ \\
\texttt{oregon1\_010526}& $\mathbf{0.24}$ & $0.36$ & $0.30$ & $2.29$ & $0.39$ & $1.47$ \\
\texttt{ca-GrQc}        & $0.32$ & $\mathbf{0.31}$ & $0.63$ & $2.17$ & $0.24$ & $1.18$ \\
\texttt{p2p-Gnutella25} & $0.38$ & $\mathbf{0.36}$ & $0.54$ & $2.33$ & $0.37$ & $1.26$ \\
\texttt{p2p-Gnutella08} & $0.44$ & $\mathbf{0.26}$ & $0.74$ & $1.97$ & $0.24$ & $1.08$ \\
\texttt{ca-HepTh}       & $0.45$ & $\mathbf{0.21}$ & $0.66$ & $2.16$ & $0.20$ & $1.13$ \\
\texttt{ca-CondMat}     & $0.82$ & $\mathbf{0.22}$ & $0.91$ & $1.81$ & $0.07$ & $0.92$ \\
\midrule
median                  & $0.38$ & $\mathbf{0.31}$ & $0.63$ & $2.17$ & $0.24$ & $1.18$ \\
\bottomrule
\end{tabular}}
\end{table}

The result parallels Section~\ref{sec:baselines} and we report it the same way. Our
fit beats both model-free estimators on every network and the graphon by a factor of
five, but the adapted inverse estimator of \cite{zhang2015estimating} is better
overall --- median $0.31$ against our $0.38$ at equal information, and $0.24$ with
$s$ supplied. We win on the two heaviest-tailed networks and lose on the four with
lighter tails, which is the same boundary that governs the exponent comparison and
the same one Proposition~\ref{prop:trans} draws.

Three remarks keep this from being read the wrong way. The gap is not an artefact of
comparing a sampled graph against expected counts: replacing our sampled realisation
by the model's closed-form expected counts \eqref{eq:Nk} at full size makes our
numbers \emph{worse}, not better (median $0.62$), because the analytic tail term
places more mass at high degree than a finite realisation does. The two model-free
estimators fail in opposite directions --- the sample understates every degree since
all are thinned, while $d_s/s$ overcorrects, mapping the many sampled degrees of $1$
to degrees of $10$, which the log scale penalises; that the obvious correction is
worse than doing nothing, and needs $s$ besides, is worth stating plainly. And a
practitioner who wants only the degree distribution should use
\cite{zhang2015estimating}: as with the edge count, the fitted graphex is justified
by returning one object that supplies the exponent, the distribution and a generator
at once, not by winning either task outright.

\subsection{Real networks: generation, and the ceiling}
\label{sec:gen}

The generative half is secondary to the estimation results and we present it as
such: what it is for, what it demonstrates, and where it stops. We train the flow on
sub-samples of a single real network and generate at full size, for seven networks
and two fitting fractions. The full tables, the anchoring analysis and the
per-network figures are in Appendix~\ref{app:gen}; two findings belong here.

\emph{The ceiling of Proposition~\ref{prop:trans} is reached, exactly.} On the three
collaboration networks, whose transitivities are $0.63$, $0.28$ and $0.26$, both our
model and the graphon baseline return values near $0.003$--$0.017$. On the
low-clustering infrastructure and peer-to-peer networks our generated transitivity is
within a factor of $2$--$3$ of the truth while the graphon baseline is $7$--$9\times$
off. \textbf{The applicability of the method is determined by whether the target's
higher-order structure is compatible with exchangeability:} infrastructure and
peer-to-peer networks are inside the class, clustering-driven collaboration networks
are not. This is the clearest empirical statement of the boundary in the paper, and
it is a prediction of the theory rather than an observation about our
implementation. Assortativity does not favour us: the AS networks have strongly
negative assortativity ($-0.17$ to $-0.19$) and, while both models recover the sign,
the graphon baseline is closer in magnitude.

\emph{What the edge counts do and do not show.} At $s=0.10$ the method leads the
graphon baseline on every network by factors of $8$--$40$ on edge count and
$3$--$17$ on degree distribution, but the advantage is a large-ratio phenomenon ---
at $s=0.25$ it narrows to $3$--$8\times$ and on \texttt{ca-CondMat} reverses --- and
the generated edge count is in any case \emph{anchored} to a target computed from
the fitted exponent. The improvement is therefore attributable to the tail estimator
and the anchoring identity, not to the generator having learned the right density.
What the experiments establish is that the sampled graphs are consistent with that
target and with the degree distribution; they do not establish that the generator is
good in any sense beyond it, and Appendix~\ref{app:gen} gives the diagnostics behind
this reading.

\begin{figure}[t]
\centering
\includegraphics[width=0.8\columnwidth]{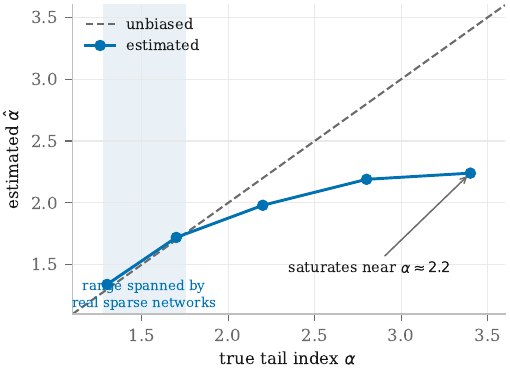}
\caption{Tail-index estimation on synthetic data with known $\alpha$. The estimator
tracks the truth inside the shaded band --- the range spanned by the fitted indices
of every genuinely sparse network in Section~\ref{sec:real} --- and saturates near
$\alpha\approx2.2$ above it. The saturation point coincides with the density at
which the method ceases to be competitive.}
\label{fig:alpha}
\end{figure}

\section{What remains open}
\label{sec:limits}

\paragraph{The clustering ceiling.} Proposition~\ref{prop:trans} gives an exact
ceiling for the \emph{rank-one} case. Our fitted, finite-capacity $K$ empirically
fails to escape that ceiling --- the capacity ablation raises $R$ sevenfold without
changing the sign of the assortativity error or moving transitivity towards the
truth --- but whether richer admissible kernels can escape it remains open. We
therefore do not claim that reaching high-clustering networks provably requires
leaving the exchangeable class; what we claim is the rank-one identity, and that
the kernels our estimator produces do not evade it.

\paragraph{The tail-index estimator saturates, and where it saturates matters.}
On synthetic data with known $\alpha$ the estimator is essentially unbiased in the
genuinely sparse range and saturates above it: it returns $1.34$, $1.72$, $1.98$,
$2.19$ and $2.24$ for true values $1.3$, $1.7$, $2.2$, $2.8$ and $3.4$
(Figure~\ref{fig:alpha}). The mechanism is informational --- a larger $\alpha$ means
a lighter tail, so separating $\alpha=2.8$ from $\alpha=3.4$ requires observations
far into the degree tail that a graph of a few hundred nodes does not contain.
Crucially, every genuinely sparse network in Section~\ref{sec:real} is fitted with
$\alpha\in[1.30,1.73]$, inside the unbiased region; the networks pushed above the
ceiling (\texttt{ca-AstroPh}, \texttt{email-Eu-core}, $\alpha\in[2.4,4.2]$) are the
dense ones where the method is not competitive anyway. The saturation point and the
scope boundary of the method coincide.

This is also where our estimator loses to the simpler one. Section~\ref{sec:baselines}
finds the graphex tail-index estimator of \cite{naulet2021bootstrap} more accurate for
edge count overall ($21.7$--$24.8\%$ against our $26.8\%$), with the deficit
concentrated in the denser networks and the two within $0.7$ percentage points inside
the sparse regime. Its estimate of $1/\sigma$ has median $1.84$ where ours has $1.62$,
and the two correlate at $0.92$ --- the same quantity, ours compressed towards the
middle of its grid. Since that estimator is a closed-form function of the degree
counts, the natural fix is not to defend our likelihood but to use theirs for the
exponent and ours for the profile; we have not pursued the combination here, and note
it as the most direct improvement available.

Four corrections were tried and none removed the saturation; we name them so they
are not re-attempted, without the numbers, which say only that each failed. A
finite-size correction fitting the analytic $n(s)$ curve is defeated by an
identifiability degeneracy between the bulk and the tail scale. A calibration map
helps below a sample-size threshold and hurts above it. A slope-continuity penalty at
the bulk/tail boundary makes the saturation monotonically worse, which at least rules
out under-determination there as the cause. Replacing the grid argmax by a
likelihood-weighted posterior mean improves the bias only by shrinking every estimate
towards the grid centre, after which the estimator stops tracking $\alpha$ at all.
Enlarging the optimisation budget does not help either.

\paragraph{Real networks are not exactly in the class.} The fitted index drifts
upward with the fitting fraction on most networks (e.g.\ \texttt{ca-HepTh}
$1.33\to2.01$). Repeating the sweep on synthetic graphexes, where $\alpha$ is
constant by construction, shows a drift of $+0.34$ attributable to the estimator
against $+0.65$ to $+0.86$ on real data, so roughly half is artifact and half is
genuine: real sparse networks do not have a single power-law tail. What is
estimated is an \emph{effective} exponent at a given observation scale, and we
phrase the claim accordingly.

\paragraph{Within-graph, not across graphs.} Every experiment here fits on a
$p$-sample of a graph and predicts that same graph, which is what the theory
describes. Whether the exponent transfers \emph{between} separately collected
networks of one kind is a different question, governed by how those networks differ
beyond size, and we make no claim about it.

\paragraph{The operating regime is narrow, and we state it as such.} Three of our own
findings bound where the method applies: the estimator is accurate for
$\alpha\lesssim1.7$ and saturates near $2.2$; the advantage over a direct log--log
regression is largest at large extrapolation ratios and narrows as the ratio shrinks;
and the fitted index drifts with the observation scale on real data, about half of
which we attribute to the estimator. Together with Section~\ref{sec:crawl} these
describe a method for \emph{large-ratio extrapolation of the densification law in the
genuinely sparse regime under uniform $p$-sampling}, not a general-purpose
size-generalising graph generator. We think the former is a real and useful thing to
have, and it is what the evidence supports.

\paragraph{The generative half rests on synthetic evidence.} The real-network
experiments measure the scaling law and the estimator, not the generator. What
supports the flow is the synthetic study of Section~\ref{sec:exp} --- it reproduces a
population of sparsity exponents including its dispersion, and Section~\ref{sec:thetabase}
shows it beats the cheaper $\theta$-space densities on validity and on distance to a
held-out population --- together with the path analysis of Section~\ref{sec:theory},
which fixes the interpolation rather than leaving it to be chosen. Section~\ref{sec:gen} adds the caveat that anchoring
\emph{imposes} the edge target, so improved generated edge counts are not evidence
that the generator learned the right density, and that after anchoring every
generated graph still has fewer edges than its target ($0.34$--$0.81$ of the true
count). Establishing a measurable benefit of the flow over simpler parameter-space
models on real corpora is open.

\paragraph{The structural comparison against graphon is not a fair test.} Our graphon
baseline is a rank-one step function, so its near-zero clustering follows by
construction. The edge-count comparison is fair --- the $\Theta(N^2)$ growth is an
identity for any fixed graphon --- but we do not read the structural columns of
Table~\ref{tab:gen} as evidence against graphon methods in general.

\section{Conclusion}

Estimating the total edge count of a partially observed sparse graph is, up to an
algebraic identity, the problem of recovering the sampling rate that produced the
observation; under a graphex the two are linked by the tail index, which makes
$e_s/s^2$ usable without knowing $s$ and explains why the quadratic graphon
estimator --- the same expression with the isolate correction omitted --- fails by a
factor rather than by a fit. The useful question is then not whether to model but
which estimator to use for which quantity, and our experiments answer it against us:
a published closed-form estimator is ahead for edge count and an adapted inverse
estimator is ahead for the degree distribution, both far behind an oracle that knows
the rate, and all of them collapsing when the sample is a crawl rather than a uniform
one. What the fit supplies that none of them does is one object carrying the
exponent, the distribution at any size and a generator, in a parameterisation where
sparsity is a coordinate rather than an emergent property --- and over that
parameterisation the interpolation path is not free, but must be geometric or the
quantity carrying sparsity is absent from the trajectory. One ceiling we can state exactly:
within the rank-one class the clustering coefficient is fixed by the degree profile,
so networks defined by their clustering lie beyond this construction and any other in
the same class.

\bibliographystyle{IEEEtran}
\bibliography{tkde_refs}

\appendices

\section{Proofs}
\label{app:proofs}

\subsection{Proposition~\ref{prop:scaling}}

The edge count is immediate: the expected number of retained pairs at size $\nu$ is
$\nu^2/2$ and each is an edge with probability $W$, so
$e_\nu=\tfrac{\nu^2}{2}\iint W$, finite by assumption.

For the node count, a point at $x$ has degree $\mathrm{Poisson}(\nu\mu(x))$ and is
non-isolated with probability $1-e^{-\nu\mu(x)}$, so
$n_\nu=\nu\int_0^\infty(1-e^{-\nu\mu(x)})\dd x$. With $\mu(x)\sim cx^{-\alpha}$
substitute $u=\nu c\,x^{-\alpha}$, giving
\begin{align}
n_\nu&=\frac{\nu(\nu c)^{1/\alpha}}{\alpha}
        \int_0^\infty(1-e^{-u})u^{-1/\alpha-1}\dd u\nonumber\\
     &=\nu(\nu c)^{1/\alpha}\Gamma\!\left(1-\tfrac1\alpha\right),
\end{align}
using $\int_0^\infty(1-e^{-u})u^{-s-1}\dd u=\Gamma(1-s)/s$ for $0<s<1$, i.e.\
$\alpha>1$. Hence $n_\nu\propto\nu^{1+1/\alpha}$, $e_\nu\propto\nu^2$, and
$e\asymp n^{2\alpha/(\alpha+1)}$.

\subsection{Proposition~\ref{prop:geom}}

For (i), $a^{1-t}b^{t}\le(1-t)a+tb\le\max(a,b)$ by weighted AM--GM, so
$W_t\in[0,1]$ and $\iint W_t\le\max(\iint W_0,\iint W_1)<\infty$. For (ii),
\begin{equation}
\psi_0^{1-t}\psi_1^{t}=c_0^{1-t}c_1^{t}(1+x)^{-(1-t)\alpha_0-t\alpha_1},
\end{equation}
again a member of the family. (iii) follows from multiplicativity of
\eqref{eq:decomp} under exponentiation, and (iv) by differentiating
$W_t=\exp((1-t)\log W_0+t\log W_1)$.

\subsection{Proposition~\ref{prop:trans}}

Edges are independent, so the expectation of a product of three distinct edge
indicators factorises exactly and the expected number of triangles is
\begin{align}
\tfrac16\sum_{i\ne j\ne k}p_{ij}p_{jk}p_{ki}
&=\tfrac16\sum_{i\ne j\ne k}\psi_i^2\psi_j^2\psi_k^2\nonumber\\
&\approx \tfrac16 S_2^3 ,
\end{align}
each triangle being counted six times among ordered triples, and the final step
drops the coincidence terms $i=j$, $j=k$, $k=i$, which contribute $O(S_2^{2}\sum\psi^4)$
against $O(S_2^{3})$. The expected degree of $j$ is $\psi_jS$, so the expected number
of connected triples is $\tfrac12\sum_j(\psi_jS)^2=\tfrac12S^2S_2$ up to the same
order. Transitivity is three times the ratio,
giving $(S_2/S)^2$. Since the points form a Poisson process of unit intensity on
$[0,\nu]\times\Rp$, $S\approx\nu\int\psi$ and $S_2\approx\nu\int\psi^2$, so $\nu$
cancels and the limit is $(\int\psi^2/\int\psi)^2$.

\subsection{Proposition~\ref{prop:conc}}

Write $\eta$ for the unit-rate Poisson process on $[0,\nu]\times\Rp$ and condition on
it; given $\eta$ the edges are independent Bernoulli variables with
$p_{ij}=W(\vartheta_i,\vartheta_j)$. We use the law of total variance twice, together
with the Poincar\'e inequality for Poisson functionals,
$\mathrm{Var}(F)\le\mathbb{E}\!\int(D_xF)^2\lambda(\dd x)$, where
$D_xF=F(\eta+\delta_x)-F(\eta)$ and $\lambda(\dd x)=\dd\theta\,\dd\vartheta$.

Throughout, $S(\vartheta):=\sum_j W(\vartheta,\vartheta_j)$ is the conditional
expected degree of a point placed at $\vartheta$. By Campbell's formula
$\mathbb{E}S=\nu\mu(\vartheta)$ and
$\mathrm{Var}(S)=\nu\!\int\!W(\vartheta,y)^2\dd y\le\nu\mu(\vartheta)$ since
$W\le1$, so
\begin{equation}
\mathbb{E}[S(\vartheta)^2]\le\nu\mu(\vartheta)+\nu^2\mu(\vartheta)^2 .
\label{eq:S2}
\end{equation}

\emph{Edges.} $\mathrm{Var}(e_\nu\mid\eta)=\sum_{i<j}p_{ij}(1-p_{ij})\le
\mathbb{E}[e_\nu\mid\eta]$, so the first term of the total-variance decomposition is
at most $e_\nu$. For the second, $F(\eta)=\mathbb{E}[e_\nu\mid\eta]=\sum_{i<j}
W(\vartheta_i,\vartheta_j)$ has $D_xF=S(\vartheta)$, so by \eqref{eq:S2}
$\mathbb{E}\!\int(D_xF)^2\lambda(\dd x)\le\nu\!\int(\nu\mu+\nu^2\mu^2)\dd\vartheta
=\nu^2\!\int\!\mu+\nu^3\!\int\!\mu^2=2e_\nu+\nu^3\!\int\!\mu^2$, using
$e_\nu=\tfrac{\nu^2}{2}\int\mu$. Adding the two gives \eqref{eq:vare}.

\emph{Vertices.} Conditionally on $\eta$, $n_\nu$ is a function of the independent
edge indicators, and flipping one changes it by at most $2$. Efron--Stein gives
$\mathrm{Var}(n_\nu\mid\eta)\le\tfrac12\sum_{i<j}4\cdot2p_{ij}(1-p_{ij})
\le4\mathbb{E}[e_\nu\mid\eta]$, so the first term is at most $4e_\nu$. For the second,
$G(\eta)=\mathbb{E}[n_\nu\mid\eta]=\sum_i\bigl(1-\prod_{j\ne i}(1-W(\vartheta_i,
\vartheta_j))\bigr)$. Adding a point at $\vartheta$ contributes its own
non-isolation probability, which is at most $S(\vartheta)$ by a union bound, and
converts previously isolated vertices, contributing at most $\sum_iW(\vartheta_i,
\vartheta)=S(\vartheta)$; both contributions are non-negative, so $0\le D_xG\le
2S(\vartheta)$ and $\mathbb{E}[(D_xG)^2]\le4\mathbb{E}[S^2]$. Integrating as before
gives $\mathrm{Var}(G)\le8e_\nu+4\nu^3\int\mu^2$, and \eqref{eq:varn} follows.

This is where the bound would otherwise diverge: a crude bound $D_xG\le1+\deg(x)$
leaves a constant to be integrated over $\vartheta\in\Rp$, and $\int_0^\infty1\,
\dd\vartheta=\infty$. Points far out in $\vartheta$ are isolated with probability
tending to one and change $n_\nu$ not at all, which is what $S(\vartheta)$ records.

\emph{Rates.} With $\mu(x)\sim cx^{-\alpha}$, $\alpha>1$, we have $\int\mu^2<\infty$,
$e_\nu=\Theta(\nu^2)$ and, by Proposition~\ref{prop:scaling},
$\mathbb{E}[n_\nu]=\Theta(\nu^{1+1/\alpha})$. Both right-hand sides in
\eqref{eq:vare}--\eqref{eq:varn} are $O(\nu^3)$, so
$\mathrm{sd}(e_\nu)/\mathbb{E}[e_\nu]=O(\nu^{3/2}/\nu^{2})$ and
$\mathrm{sd}(n_\nu)/\mathbb{E}[n_\nu]=O(\nu^{3/2}/\nu^{1+1/\alpha})$, which is
\eqref{eq:rates}. The latter tends to zero exactly when $1/\alpha>1/2$.

\subsection{Proposition~\ref{prop:remainder}}

Write $N(\nu)=\mathbb{E}[n_\nu]/\nu=\int_0^\infty(1-e^{-\nu\mu(x)})\dd x$ and
$g(x)=cx^{-\alpha}$. Split at $x_1\ge x_0$ chosen so that
$Cx^{-\rho}\le\tfrac12 c$ for $x\ge x_1$, which makes $\mu\ge g/2$ there.

On $[0,x_1]$ the integrand is bounded by $1$, contributing $O(1)$; against the main
term $\Theta(\nu^{1/\alpha})$ this is $O(\nu^{-1/\alpha})$, the second error term in
\eqref{eq:remrate}.

On $(x_1,\infty)$ we compare with $g$. The crude bound $|e^{-a}-e^{-b}|\le|a-b|$
gives $\nu C\int x^{-\alpha-\rho}\dd x=O(\nu)$, which exceeds the main term for
$\alpha>1$ and is therefore useless; we use instead
$|e^{-a}-e^{-b}|\le|a-b|e^{-\min(a,b)}$ with $\min(\nu g,\nu\mu)\ge\nu g/2$, giving
\begin{equation*}
\int_{x_1}^{\infty}\!\!\bigl|e^{-\nu g}-e^{-\nu\mu}\bigr|\dd x
\;\le\;\nu C\!\int_{x_1}^{\infty}\!\! x^{-\alpha-\rho}e^{-\nu cx^{-\alpha}/2}\dd x .
\end{equation*}
Substituting $t=\nu cx^{-\alpha}/2$, so $x=(\nu c/2t)^{1/\alpha}$ and
$\dd x=-\tfrac1\alpha(\nu c/2)^{1/\alpha}t^{-1/\alpha-1}\dd t$, the right-hand side is
\begin{equation*}
\frac{\nu C}{\alpha}\Bigl(\frac{\nu c}{2}\Bigr)^{\!1/\alpha}
\Bigl(\frac{2}{\nu c}\Bigr)^{\!1+\rho/\alpha}
\int_0^{\infty}\!\! t^{\,\rho/\alpha-1/\alpha}e^{-t}\dd t
\;=\;O\!\left(\nu^{(1-\rho)/\alpha}\right),
\end{equation*}
the Gamma integral converging because $\alpha+\rho>1$. Relative to the main term
$\Theta(\nu^{1/\alpha})$ this is $O(\nu^{-\rho/\alpha})$. Finally
$\int_{x_1}^\infty(1-e^{-\nu g})\dd x=(\nu c)^{1/\alpha}\Gamma(1-1/\alpha)+O(1)$ by
the exact computation above, which gives \eqref{eq:remrate}.

For the ratio, apply \eqref{eq:remrate} at $s\nu$ and at $\nu$ and divide; the
$\nu$-term errors are dominated by the $s\nu$-term ones, leaving
$s^{1+1/\alpha}(1+O((s\nu)^{-\rho/\alpha})+O((s\nu)^{-1/\alpha}))$, i.e.\ the stated
$\tau=\min(\rho,1)/\alpha$.

\subsection{The tail term of \eqref{eq:Nk}}

On the tail $\mu(u)=\mu_R e^{-\alpha(u-U_R)}$ with $\dd x=e^{u}\dd u$; writing
$v=u-U_R$ and $w=\mu_R e^{-\alpha v}$,
\begin{align}
&\int_{U_R}^{\infty}\mathrm{Poisson}(k;\mu(u))e^{u}\dd u\nonumber\\
&\quad=\frac{e^{U_R}\mu_R^{1/\alpha}}{\alpha\,k!}
        \int_0^{\mu_R}e^{-w}w^{\,k-1/\alpha-1}\dd w\nonumber\\
&\quad=\frac{e^{U_R}\mu_R^{1/\alpha}}{\alpha\,k!}
        \gamma\!\left(k-\tfrac1\alpha,\mu_R\right),
\end{align}
convergent at $w=0$ since $k-1/\alpha-1>-1$ for $k\ge1$, $\alpha>1$.

\section{Generation on real networks}
\label{app:gen}

\begin{figure*}[t]
\centering
\includegraphics[width=0.44\textwidth]{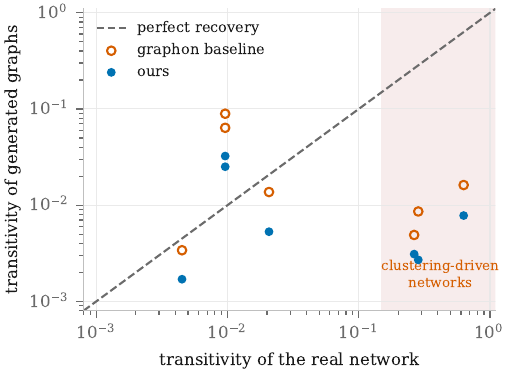}
\hfill
\includegraphics[width=0.44\textwidth]{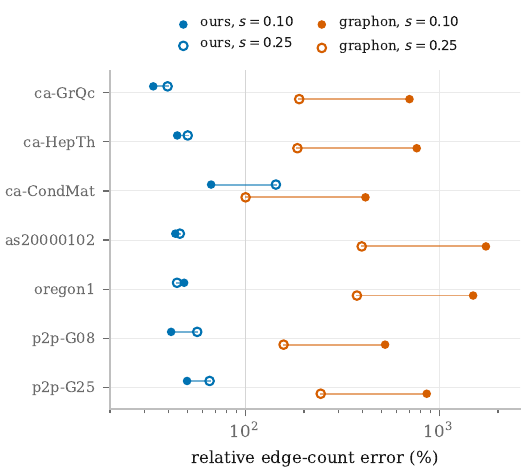}
\caption{\textbf{Left:} transitivity of generated graphs against that of the target
network. Both models saturate near $10^{-2}$ regardless of the target. For the
rank-one case Proposition~\ref{prop:trans} says exactly why; for our fitted $K$ this
is an empirical observation that the ceiling is not escaped, not a proof that it
cannot be. \textbf{Right:} edge-count error at two
fitting fractions. Our error is nearly flat in the fraction, while the graphon
baseline improves steeply as the extrapolation ratio shrinks --- so the advantage
is a large-ratio phenomenon, and at $s=0.25$ on \texttt{ca-CondMat} it reverses.}
\label{fig:gen}
\end{figure*}

We next train the flow on sub-samples of a single real network and generate at the
full size, for seven networks and two fitting fractions. Table~\ref{tab:gen}
reports the outcome at $s=0.10$ and Figure~\ref{fig:gen} both fractions.

\begin{table}[t]
\caption{Training on $10\%$ sub-samples and generating at full size (a $10$--$30\times$
extrapolation). The model recovers edge count and degree distribution; clustering
behaves exactly as Proposition~\ref{prop:trans} predicts. \texttt{as20000102} and
\texttt{oregon1} are low-clustering networks and are reproduced within the class.}
\label{tab:gen}
\centering
\footnotesize
\resizebox{\columnwidth}{!}{%
\begin{tabular}{lccccccc}
\toprule
& \multicolumn{2}{c}{transitivity} & \multicolumn{2}{c}{degW1} &
\multicolumn{2}{c}{edge error} & \\
\cmidrule(lr){2-3}\cmidrule(lr){4-5}\cmidrule(lr){6-7}
network & real & ours & ours & graphon & ours & graphon & \\
\midrule
\texttt{ca-GrQc}        & $0.630$ & $0.008$ & $\mathbf{0.32}$ & $2.17$ & $\mathbf{33\%}$ & $700\%$ & \\
\texttt{ca-HepTh}       & $0.284$ & $0.003$ & $\mathbf{0.45}$ & $2.16$ & $\mathbf{44\%}$ & $763\%$ & \\
\texttt{ca-CondMat}     & $0.264$ & $0.003$ & $\mathbf{0.82}$ & $1.81$ & $\mathbf{66\%}$ & $415\%$ & \\
\texttt{p2p-Gnutella08} & $0.021$ & $0.005$ & $\mathbf{0.44}$ & $1.97$ & $\mathbf{41\%}$ & $524\%$ & \\
\texttt{as20000102}     & $0.010$ & $\mathbf{0.032}$ & $\mathbf{0.17}$ & $2.83$ & $\mathbf{43\%}$ & $1735\%$ & \\
\texttt{oregon1\_010526}& $0.010$ & $\mathbf{0.025}$ & $\mathbf{0.24}$ & $2.29$ & $\mathbf{48\%}$ & $1490\%$ & \\
\texttt{p2p-Gnutella25} & $0.005$ & $0.002$ & $\mathbf{0.38}$ & $2.33$ & $\mathbf{50\%}$ & $859\%$ & \\
\bottomrule
\end{tabular}}
\end{table}

At $s=0.10$ the method leads on every network, by factors of $8$ to $40$ on edge
count and $3$ to $17$ on degree distribution. Three qualifications matter.

\paragraph{The advantage is a large-ratio phenomenon.} The graphon error grows like
$R^{2-\beta}$, so it is only severe when the extrapolation ratio $R$ is large. At
$s=0.25$ ($R\approx4$) our margin narrows to $3$--$8\times$ and on
\texttt{ca-CondMat} it reverses outright ($143\%$ against $100\%$). A reader should
take the method as addressing large-ratio densification extrapolation, not as
uniformly better graph generation.

\paragraph{Naive generation is unstable in the fitting fraction, and we can say
why.} Solving $\nu$ directly for the target node count, the two largest networks
under-produce edges at $s=0.10$ (\texttt{ca-CondMat} $66\%$ low,
\texttt{p2p-Gnutella25} $46\%$ low) and \emph{over}-produce at $s=0.25$ ($248\%$ and
$85\%$ high). Two diagnostics locate the cause. First, the analytic edge count
$\tfrac{\nu^2}{2}(\int\psi)^2$ agrees with the sampled count to within $6\%$ in
every case and the sampler's point-count guard never triggers, so the error is
already present in the fitted parameters rather than introduced by sampling.
Second, the two networks that over-produce are exactly the two whose fitted tail
index rises sharply with the fitting fraction ($1.40\to1.94$ and $1.40\to1.62$),
while the stable networks have a flat or falling index. This is
Proposition~\ref{prop:prop} evaluated at $R=n_{\mathrm{full}}$: with
$\ln n_{\mathrm{full}}\approx10$, a $\Delta\beta$ of $0.15$ is already a factor
$4.7$.

\paragraph{Anchoring removes the pathology.} Table~\ref{tab:anchor} compares the two
generation rules. Anchoring eliminates over-production entirely --- no generated
graph exceeds the true edge count --- and cuts the two pathological errors from
$248\%$ and $85\%$ to $42\%$ and $29\%$. The median error over $14$ settings falls
from $48\%$ to $39\%$, with anchoring better in $10$. A residual remains and is
accounted for by the same tail-index bias: every anchored ratio is below one
($0.34$--$0.81$), which is what $R^{\hat\beta-\beta}$ gives for
$\Delta\beta\approx-0.1$ at $\ln R\approx3$. Anchoring therefore does what it was
designed to do --- it reduces the effective extrapolation ratio from
$n_{\mathrm{full}}$ to $R$ --- and leaves the estimator bias acting through the
smaller ratio, bringing generation close to the accuracy of the extrapolation
formula itself ($39\%$ against $26.8\%$).

We separate two claims that this table could be read as conflating. Anchoring
\emph{imposes} the edge target $e_sR^{\beta}$, so the improvement in generated edge
counts is attributable to the tail estimator and the anchoring identity, not to the
generative model having learned the right density; a graph generator that merely
respects an imposed target is not thereby validated. What the generation experiments
do establish is that the sampled graphs are consistent with that target and with the
degree distribution; what they do not establish is that the generator is good in any
sense beyond it. \textbf{The scaling predictor is what our evidence supports; the
graph generator is supported much more weakly.}

\begin{table}[t]
\caption{Naive against anchored generation. The two over-producing settings are in
bold; anchoring removes over-production everywhere.}
\label{tab:anchor}
\centering
\footnotesize
\resizebox{\columnwidth}{!}{%
\begin{tabular}{lcccccc}
\toprule
& & & \multicolumn{2}{c}{$e_{\mathrm{gen}}/e_{\mathrm{real}}$} &
\multicolumn{2}{c}{relative error} \\
\cmidrule(lr){4-5}\cmidrule(lr){6-7}
network & $s$ & $\hat\alpha$ & naive & anchored & naive & anchored \\
\midrule
\texttt{ca-CondMat}     & $0.25$ & $1.94$ & $\mathbf{3.48}$ & $0.58$ & $248\%$ & $\mathbf{42\%}$ \\
\texttt{p2p-Gnutella25} & $0.25$ & $1.62$ & $\mathbf{1.85}$ & $0.71$ & $85\%$  & $\mathbf{29\%}$ \\
\texttt{as20000102}     & $0.10$ & $1.40$ & $0.48$ & $0.81$ & $52\%$  & $\mathbf{19\%}$ \\
\texttt{p2p-Gnutella08} & $0.25$ & $1.51$ & $0.43$ & $0.58$ & $57\%$  & $\mathbf{42\%}$ \\
\texttt{oregon1\_010526}& $0.25$ & $1.30$ & $0.57$ & $0.64$ & $43\%$  & $\mathbf{36\%}$ \\
\texttt{ca-GrQc}        & $0.25$ & $1.62$ & $0.63$ & $0.66$ & $37\%$  & $\mathbf{34\%}$ \\
\texttt{ca-HepTh}       & $0.25$ & $1.40$ & $0.52$ & $0.48$ & $\mathbf{48\%}$ & $52\%$ \\
\midrule
\multicolumn{3}{l}{median over all $14$ settings} & & & $48\%$ & $\mathbf{39\%}$ \\
\bottomrule
\end{tabular}}
\end{table}

\paragraph{Clustering behaves exactly as predicted, and assortativity does not
favour us.} On the three collaboration networks, whose transitivities are $0.63$,
$0.28$ and $0.26$, both models return values near $0.003$--$0.017$: the ceiling of
Proposition~\ref{prop:trans}. On the low-clustering networks our generated
transitivity is within a factor of $2$--$3$ of the truth while the graphon baseline
is $7$--$9\times$ off. Assortativity, however, goes the other way: the AS networks
have strongly negative assortativity ($-0.17$ to $-0.19$), and while both models
recover the sign, the graphon baseline is closer in magnitude ($-0.06$ to $-0.09$)
than ours ($-0.015$ to $-0.049$). \textbf{The applicability of the method is
determined by whether the target's higher-order structure is compatible with
exchangeability:} infrastructure and peer-to-peer networks are inside the class,
clustering-driven collaboration networks are not.

\section{Experimental details}
\label{app:details}

\paragraph{Synthetic graphexes.} Tail indices are drawn uniformly from
$[1.25,2.6]$; the bulk profile is monotone decreasing with a random smooth
perturbation; $\kappa$ is a two-block assortative structure with random strength,
constrained non-positive so $\ell\le0$ holds automatically.

\paragraph{Flow.} Four-layer MLP of width $512$ with sinusoidal time embedding,
Adam at $2\times10^{-3}$ with cosine decay, $40\,000$ steps, batch $256$; sampling
uses $100$ Euler steps. Three seeds; dispersion reported in Table~\ref{tab:round}.

\paragraph{Estimator.} The tail index is gridded over $30$ values in $[1.08,4.2]$
with $260$ Adam steps per value; see Section~\ref{sec:limits} for why a larger
budget does not help.

\paragraph{Sampling.} Bulk--bulk pairs are drawn exactly, in chunks to bound
memory; pairs involving the tail are rank-one and drawn by a Poisson approximation,
exact to leading order there since $\psi_i\psi_j\ll1$.

\paragraph{Real networks.} Undirected simple graphs from SNAP edge lists, obtained
by symmetrising, removing self-loops and de-duplicating. Sub-sampling retains nodes
independently and takes the induced subgraph, \textbf{then discards vertices that
are isolated in it}. This is the standard graphex $p$-sampling convention and is the
one consistent with the theory: $n_\nu=\nu\int(1-e^{-\nu\mu})\dd x$ counts
non-isolated vertices only, so an observed graph must be the non-isolated part. It
also matters numerically --- retaining new isolates would change the $(n,e)$ scaling
of every sub-sample and hence every exponent fitted from one --- so we state it
explicitly rather than leaving it to the reader. The full list is
\texttt{as20000102}, \texttt{oregon1\_010331}, \texttt{oregon1\_010526},
\texttt{p2p-Gnutella04/08/25/31}, \texttt{ca-GrQc}, \texttt{ca-HepTh},
\texttt{ca-CondMat}, \texttt{ca-HepPh}, \texttt{ca-AstroPh},
\texttt{email-Eu-core}.

\end{document}